\documentclass[a4paper]{template/svproc}
\usepackage{url}

\usepackage{bmpsize}
\usepackage[linesnumbered,ruled,vlined]{algorithm2e}
\usepackage{amssymb,enumerate}
\usepackage{mathrsfs}
\usepackage{times}
\usepackage{multicol}
\usepackage{amsfonts}
\usepackage{textcomp}
\usepackage{balance}
\usepackage{multirow}
\usepackage{booktabs}
\usepackage{dblfloatfix} 
\usepackage{bbold}
\usepackage{makecell}
\usepackage{hhline} 
\usepackage{float} 
\usepackage{xparse}
\usepackage{lipsum}
\usepackage{soul} 
\usepackage{times}
\usepackage{cite}
\usepackage{balance}
\usepackage{dsfont} 
\usepackage{physics} 
\usepackage{setspace} 
\usepackage{accents} 

\usepackage[dvipdfmx]{graphicx} 
\usepackage[T1]{fontenc}
\usepackage[usenames,dvipsnames,svgnames,table, xcdraw]{xcolor}
\usepackage[caption=false,font=footnotesize]{subfig}
\usepackage[utf8]{inputenc}

\usepackage{hyperref}
\hypersetup{
  colorlinks=true,pdfpagemode=UseNone,citecolor=black,linkcolor=black,urlcolor=BrickRed
}

\usepackage{caption}
\SetCommentSty{mycommfont}

\SetKwInput{KwInput}{Input}                
\SetKwInput{KwOutput}{Output}              

\graphicspath{
    {graphics/}
}

\newcommand{\bh}[1]{[{\textbf{\textcolor{blue}{Bruce: #1}}}]}

\newcommand{\bht}[1]{[{\textbf{\textcolor{note}{Bruce: (TODO) #1}}}]}

\newcommand{\vrdb}[2]{{\sethlcolor{cyan}\hl{#1} \marginpar{\fbox{\parbox{75pt}{\raggedright\scriptsize\textbf{\textcolor{cyan}{Vishnu: #2}}}}}}}

\newcommand{\bhdd}[1]{[{\textbf{\textcolor{red}{Bruce: DONE. If you were to change something, use \textbackslash\texttt{add}\{\} to indicate where you changed!!}}}]}
\newcommand{\bhww}[1]{[{\textbf{\textcolor{red}{Bruce: WORKING IN PROGRESS. If you were to change something, use \textbackslash\texttt{add}\{\} to indicate where you changed!!}}}]}

\definecolor{note}{rgb}{0.3,0.7,0.25}
\definecolor{rephase}{rgb}{0.15,0.7,0.15}
\definecolor{bag}{rgb}{0.5,0.5,0.0}
\definecolor{lightgreen}{rgb}{0.1,0.8,0.1}
\definecolor{lightblack}{rgb}{0.1,0.1,0.1}

\newcommand{\bdd}[1]{\textbf{\textcolor{blue}{#1}}}

\newcommand{\Astar}{$\text{A}^*$~}

\newcommand{\imomd}{$\text{IMOMD-RRT}^*$~}
\newcommand{\imomdN}{$\text{IMOMD-RRT}^*$}

\newcommand{\tspsolver}{ECI-Gen solver~}
\newcommand{\tspsolverN}{ECI-Gen solver}
\newcommand{\imomt}{\imomd}
\newcommand{\imomtN}{\imomdN}
\newcommand{\rrt}{$\text{RRT}^*$~}
\newcommand{\rrtN}{$\text{RRT}^*$}

\newcommand{\vsub}[1]{v_{\text{#1}}}

\newcommand{\squeezeup}{\vspace{-4mm}} 

\newcommand{\comment}[1]{}
\newcommand{\conference}[1]{#1}

\DeclareDocumentCommand{\RI}{ O{} O{} }{\mathcal{RI}_{#1}^{#2}}

\DeclareDocumentCommand{\td}{ O{} }{\tilde{#1}}

\DeclareDocumentCommand{\asin}{ O{} }{\sin^{-1}(#1)}
\DeclareDocumentCommand{\acos}{ O{} }{\cos^{-1}(#1)}
\DeclareDocumentCommand{\atan}{ O{} }{\tan^{-1}(#1)}

\DeclareDocumentCommand{\vector}{ O{} }{\mathrm{vec}(#1)}
\DeclareDocumentCommand{\zeros}{ O{} }{\textbf{0}_{#1}}
\DeclareDocumentCommand{\pre}{ O{} O{} }{{}_{#1}^{#2}}

\DeclareMathOperator*{\argmin}{arg\,min}
\DeclareMathOperator*{\argmax}{arg\,max}

\newcommand{\Ccal}{\mathcal{C}}
\newcommand{\Dcal}{\mathcal{D}}
\newcommand{\Ecal}{\mathcal{E}}

\newcommand{\Gcal}{\mathcal{G}}

\newcommand{\Kcal}{\mathcal{K}}

\newcommand{\Mcal}{\mathcal{M}}
\newcommand{\Ncal}{\mathcal{N}}
\newcommand{\Ocal}{\mathcal{O}}

\newcommand{\Qcal}{\mathcal{Q}}

\newcommand{\Scal}{\mathcal{S}}
\newcommand{\Tcal}{\mathcal{T}}
\newcommand{\Vcal}{\mathcal{V}}
\newcommand{\Xcal}{\mathcal{X}}

\DeclareDocumentCommand{\A}{ O{} O{} }{\textbf{A}_{#1}^{#2}}
\DeclareDocumentCommand{\H}{ O{} O{} }{\textbf{H}_{#1}^{#2}}
\DeclareDocumentCommand{\I}{ O{} O{} }{\textbf{I}_{#1}^{#2}}
\DeclareDocumentCommand{\L}{ O{} O{} }{\textbf{L}_{#1}^{#2}}
\DeclareDocumentCommand{\M}{ O{} O{} }{\textbf{M}_{#1}^{#2}}
\DeclareDocumentCommand{\N}{ O{} O{} }{\textbf{N}_{#1}^{#2}}
\DeclareDocumentCommand{\O}{ O{} O{} }{\textbf{O}_{#1}^{#2}}
\DeclareDocumentCommand{\P}{ O{} O{} }{\textbf{P}_{#1}^{#2}}
\DeclareDocumentCommand{\Q}{ O{} O{} }{\textbf{Q}_{#1}^{#2}}
\DeclareDocumentCommand{\R}{ O{} O{} }{\textbf{R}_{#1}^{#2}}
\DeclareDocumentCommand{\T}{ O{} O{} }{\textbf{T}_{#1}^{#2}}
\DeclareDocumentCommand{\U}{ O{} O{} }{\textbf{U}_{#1}^{#2}}
\DeclareDocumentCommand{\V}{ O{} O{} }{\textbf{V}_{#1}^{#2}}
\DeclareDocumentCommand{\X}{ O{} O{} }{\textbf{X}_{#1}^{#2}}
\DeclareDocumentCommand{\Y}{ O{} O{} }{\textbf{Y}_{#1}^{#2}}
\DeclareDocumentCommand{\Z}{ O{} O{} }{\textbf{Z}_{#1}^{#2}}

\DeclareDocumentCommand{\e}{ O{} O{} }{\textbf{e}_{#1}^{#2}}
\DeclareDocumentCommand{\n}{ O{} O{} }{\textbf{n}_{#1}^{#2}}
\DeclareDocumentCommand{\o}{ O{} O{} }{\textbf{o}_{#1}^{#2}}
\DeclareDocumentCommand{\t}{ O{} O{} }{\textbf{t}_{#1}^{#2}}
\DeclareDocumentCommand{\p}{ O{} O{} }{\textbf{p}_{#1}^{#2}}
\DeclareDocumentCommand{\q}{ O{} O{} }{\textbf{q}_{#1}^{#2}}
\DeclareDocumentCommand{\r}{ O{} O{} }{\textbf{r}_{#1}^{#2}}
\DeclareDocumentCommand{\u}{ O{} O{} }{\textbf{u}_{#1}^{#2}}
\DeclareDocumentCommand{\v}{ O{} O{} }{\textbf{v}_{#1}^{#2}}
\DeclareDocumentCommand{\x}{ O{} O{} }{\textbf{x}_{#1}^{#2}}

\begin{document}
\mainmatter              
\title{Informable Multi-Objective and Multi-Directional RRT$^\ast$\\ System for Robot Path Planning}
\titlerunning{IMOMD-RRT$^\ast$ for Robot Path Planning}  
%
\author{Jiunn-Kai Huang\inst{1} \and Yingwen Tan\inst{1} \and Dongmyeong Lee\inst{1} \and \\
    Vishnu R. Desaraju\inst{2} \and Jessy W. Grizzle\inst{1}
}
%
\authorrunning{J.K. Huang et al.} 
%
%
\institute{Robotics Department, University of Michigan, Ann Arbor, MI 48109, USA,\\
\email{\{bjhuang, tywwyt, domlee, grizzle\}@umich.edu},\\ home page:
\texttt{https://www.biped.solutions/}
\and
Woven Planet North America, Ann Arbor, MI 48105, USA.
\email{vishnu.desaraju@woven-planet.global}
}
\maketitle              
\begin{abstract}
        Multi-objective or multi-destination path planning is crucial for mobile robotics
        applications such as mobility as a service, robotics inspection, and electric vehicle
    charging for long trips. This work proposes an anytime iterative system to
    concurrently solve the multi-objective path planning problem and determine the
    visiting order of destinations. The system is comprised of an anytime informable
    multi-objective and multi-directional RRT$^*$ algorithm to form a simple connected graph, and a proposed solver that consists of an enhanced cheapest insertion algorithm and a genetic algorithm
    to solve the relaxed traveling salesman problem in polynomial time. Moreover,
    a list of waypoints is often provided for robotics
    inspection and vehicle routing so that the robot can preferentially visit certain equipment or
    areas of interest. We show that the proposed system can inherently incorporate
    such knowledge, and can navigate through challenging topology. The proposed anytime system is evaluated on large and complex
    graphs built for real-world driving applications.\comment{Additionally, the proposed relaxed-TSP solver is benchmarked with the greedy search algorithm and an open-sourced TSP solver.} All implementations are coded in
    multi-threaded C++ and are available at:
    \href{https://github.com/UMich-BipedLab/IMOMD-RRTStar}{https://github.com/UMich-BipedLab/IMOMD-RRTStar}.

\end{abstract}

\section{Introduction and Contributions}
\label{sec:Intro}


Multi-objective or multi-destination path planning is a key enabler of applications
such as data collection \cite{faigl2014unifying, hu2020aoi, samir2019uav}, 
traditional Traveling Salesman Problem (TSP)
\cite{junger1995traveling,applegate2011traveling}, and electric
vehicle charging for long trips. More recently, autonomous ``Mobility as a Service''
(e.g., autonomous shuttles or other local transport between user-selected points) has
become another important application of multi-objective planning such as car-pooling
\cite{ma2018path, huang2018multimodal, al2019deeppool, hulagu2020electric} 
Therefore, being able to efficiently find paths
connecting multiple destinations and to determine the visiting order of the
destinations (essentially, a relaxed TSP) is critical for modern navigation systems
deployed by autonomous vehicles and robots. This paper seeks to solve these two
problems by developing a system composed of a sampling-based anytime path planning
algorithm and a relaxed-TSP solver.

Another application of multi-objective path planning is robotics inspection, where a list of inspection waypoints is often provided so that the robot can
preferentially examine certain equipment or areas of interest or avoid certain areas
in a factory, for example. The pre-defined waypoints can be considered as prior
knowledge for the overall inspection path the robot needs to construct for task
completion. We show that the proposed system can inherently incorporate such
knowledge.



\begin{figure}[!t]%
    \centering
    \subfloat[]{%
    \label{fig:FirstImg1}%
    \includegraphics[trim=0 0 0 0,clip,height=0.18\columnwidth]{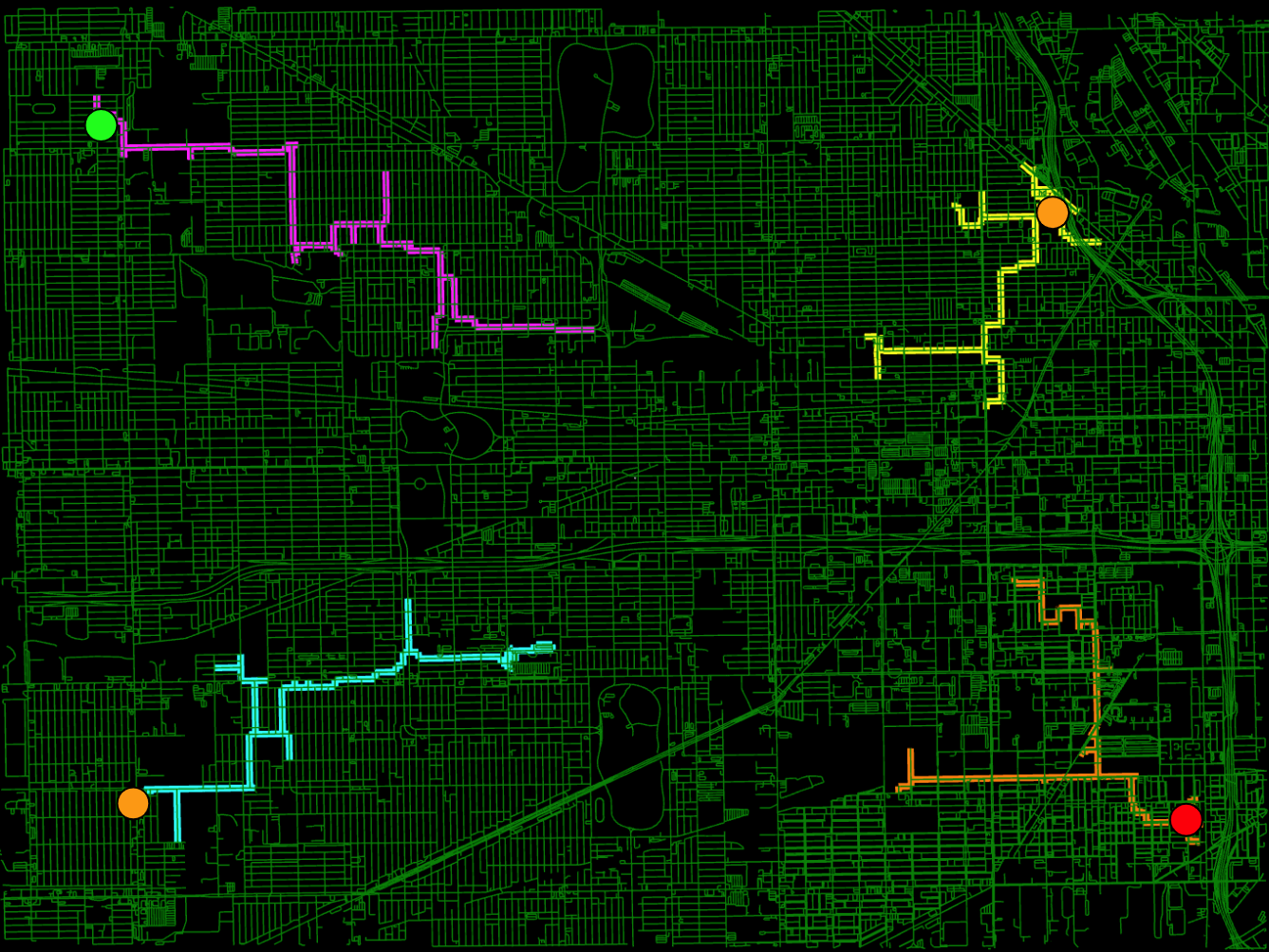}}~
    \subfloat[]{%
    \label{fig:FirstImg2}%
    \includegraphics[trim=0 0 0 0,clip,height=0.18\columnwidth]{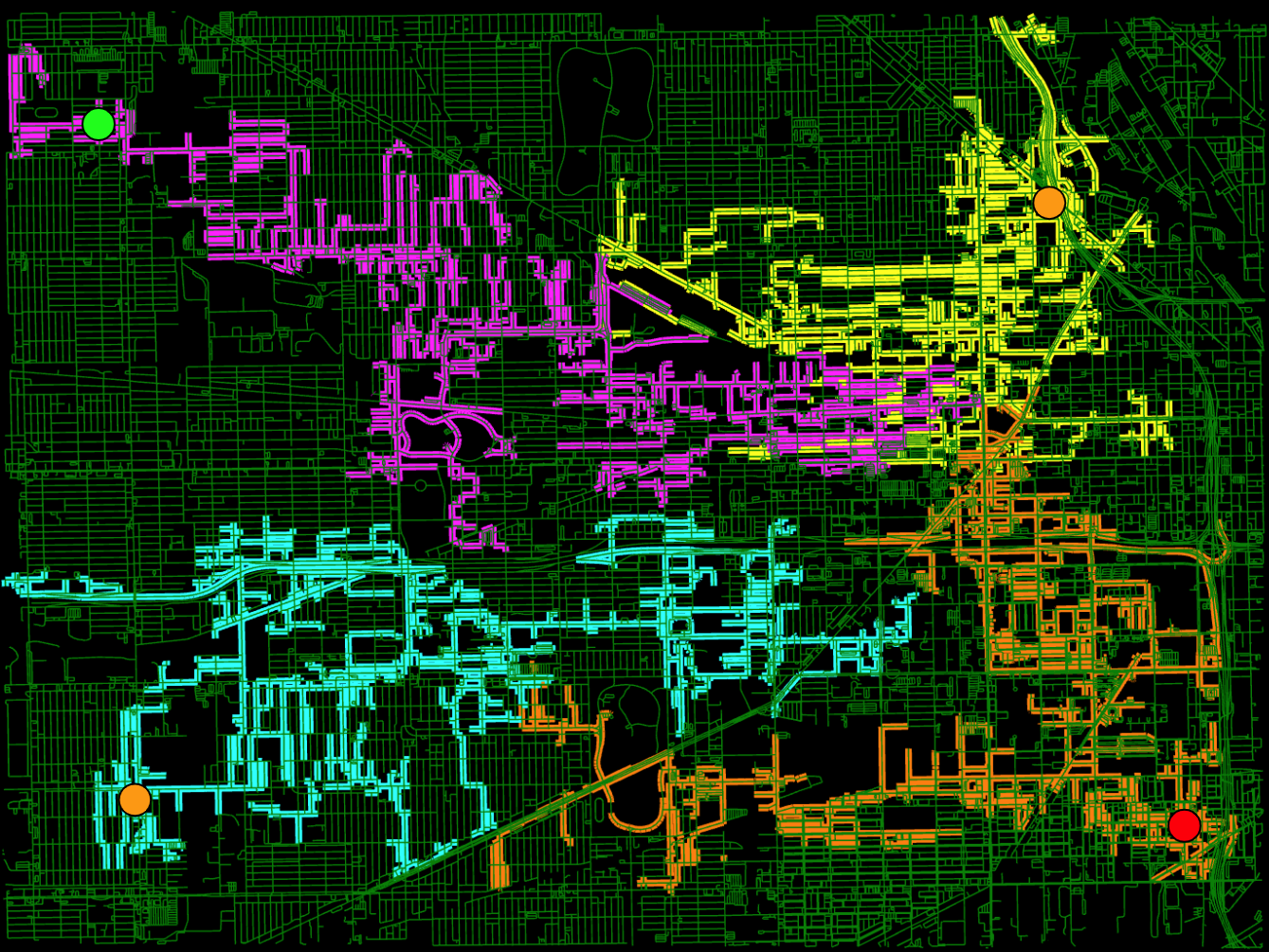}}~
    \subfloat[]{%
    \label{fig:FirstImg3}%
    \includegraphics[trim=0 0 0 0,clip,height=0.18\columnwidth]{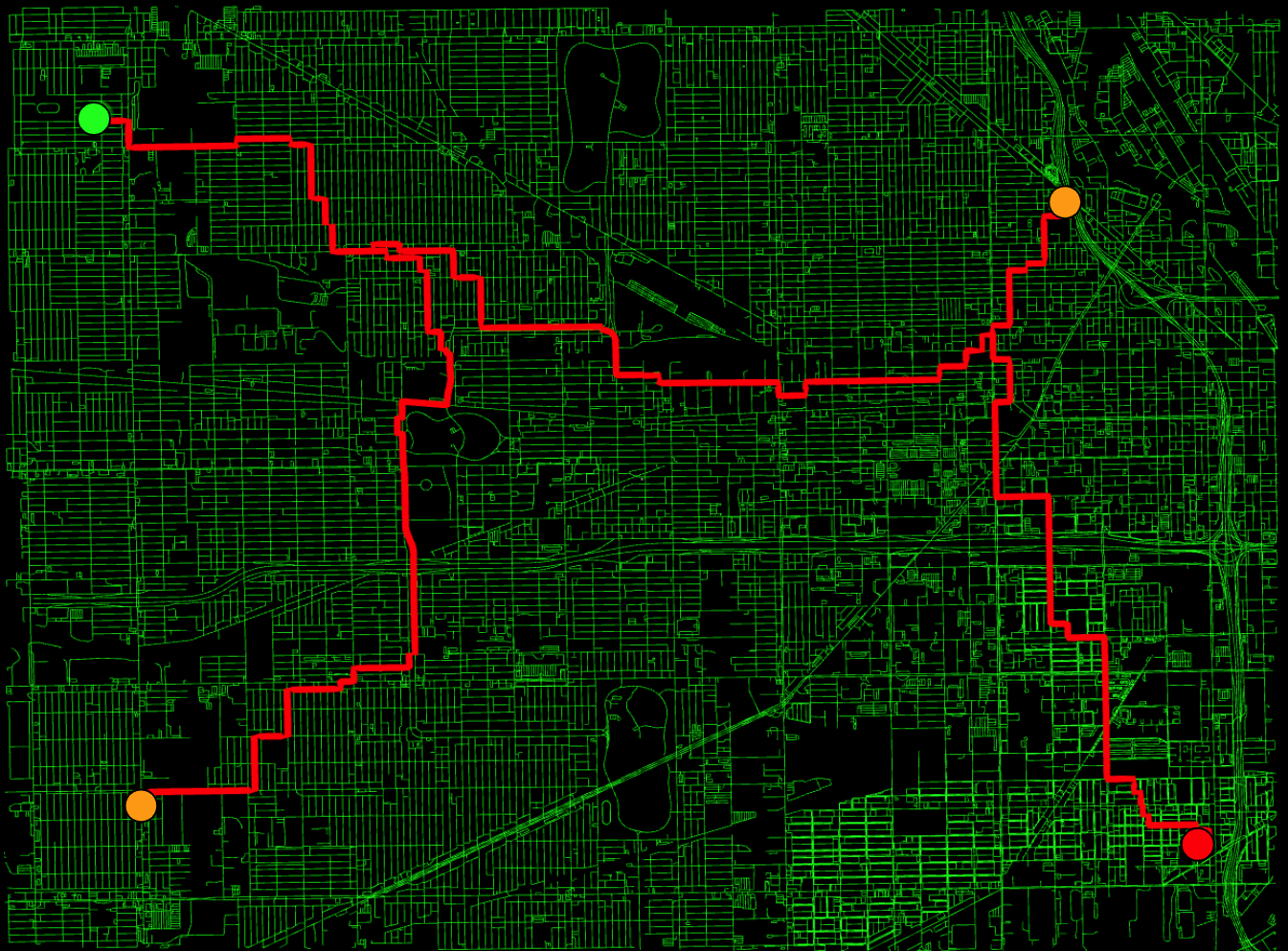}}~
    \subfloat[]{%
    \label{fig:FirstImg4}%
    \includegraphics[trim=0 0 0 0,clip,height=0.18\columnwidth]{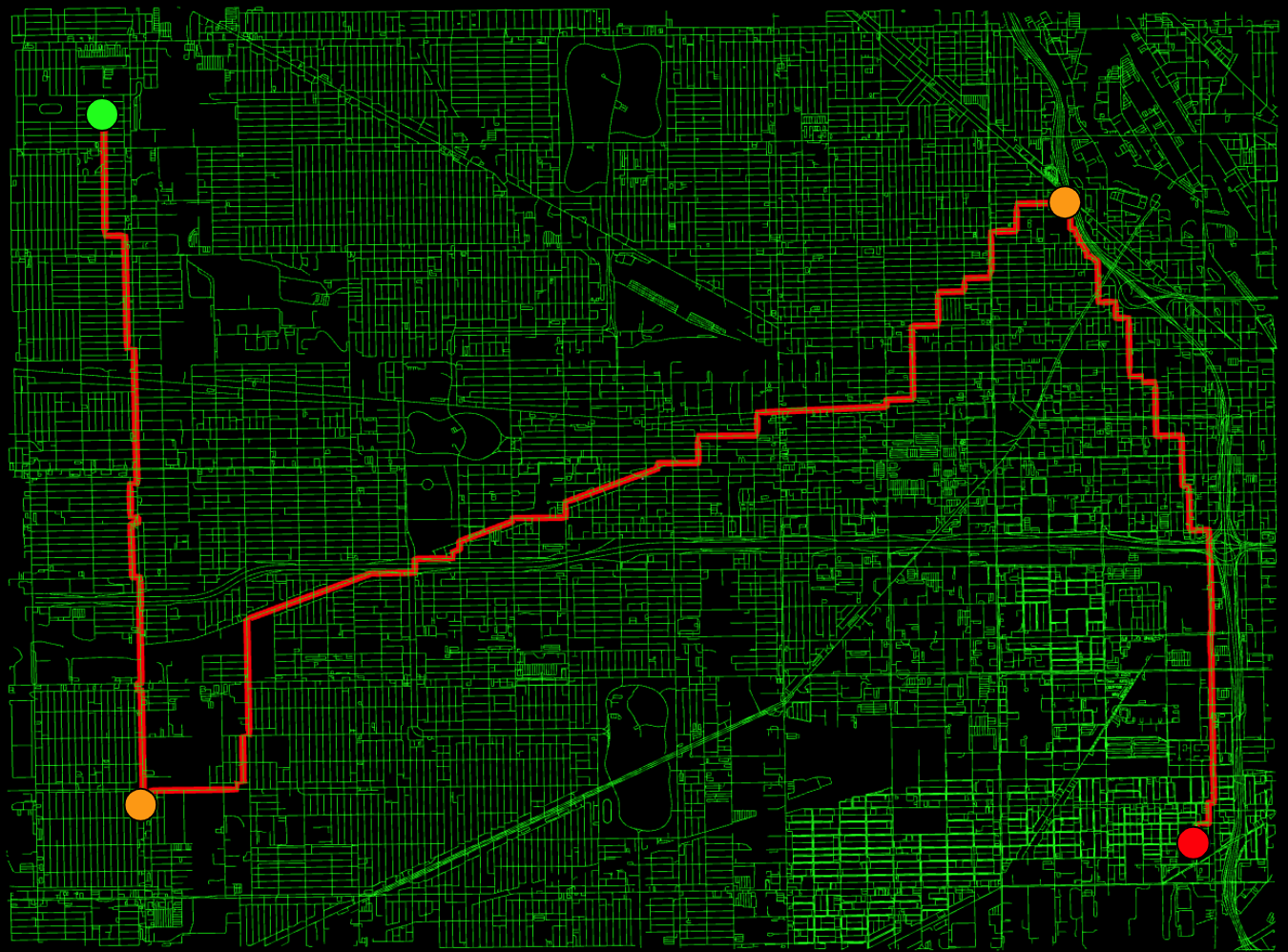}}%
    \caption[]{{Illustration of the proposed informable multi-objective and
            multi-directional RRT$^\ast$ (\imomdN) system evaluated on OpenStreetMap
            of Chicago, containing $866,089$ nodes and $1,038,414$ edges. The green,
            red, and orange dots are the source, target, and objectives,
            respectively. \subref{fig:FirstImg1} shows the initial stage of the tree
            expansion of each destination. \subref{fig:FirstImg2} shows the trees from each destination form a connected graph. \subref{fig:FirstImg3}
            shows the first path and visiting order from the \imomdN. Given more
            computation time, \subref{fig:FirstImg4} shows that \imomd returns a
            better path and order.}
    }%
    \label{fig:FirstImg}%
    \squeezeup
    \vspace{-3mm}
\end{figure}


To represent multiple destinations, graphs composed of nodes and edges stand out for
their sparse representations. In particular, graphs are a popular representation of
topological landscape features, such as terrain contour, lane markers, or
intersections \cite{OpenStreetMap}. Topological features do not change often; thus,
they are maintainable and suitable for long-term support compared to high-definition
(HD) maps. Graph-based maps such as OpenStreetMap \cite{OpenStreetMap} have been
developed over the past two decades to describe topological features and are readily
available worldwide. Therefore, we concentrate on developing the proposed informable
multi-objective and multi-directional rapidly-exploring random trees (RRT$^\ast$)
system for path planning on large and complex graphs.

A multi-objective path planning system is charged with two tasks: \textbf{1)} find weighted paths (i.e., paths and traversal costs), if they exist, that connect the various destinations. This operation
results in an undirected and weighted graph where nodes and edges correspond to destinations
and paths connecting destinations, respectively. \textbf{2)} determine the visiting
order of destinations that minimizes total travel cost. The second task, called
relaxed TSP, differs from standard TSP in that we are allowed to (or sometimes have
to) visit a node multiple times; see Sec.~\ref{sec:MO-TSP} for a detailed discussion.
Several approaches \cite{janovs2021multi, devaurs2014multi, vonasek2019space,
englot2011multi} have been developed to solve each of these tasks separately,
assuming either the visiting order of the destinations is given, or a cyclic/complete
graph is constructed and the weights of edges are provided. However, in real-time
applications, the connectivity of the destinations and the weights of the paths
between destinations (task 1) as well as the visiting order of the destinations (task
2) are often unknown. It is crucial that we are able to solve these two tasks
concurrently in an anytime manner, meaning that the system can provide suboptimal but monotonically improving
solutions at any time throughout the path cost minimization process. 


In this paper, we seek to develop an anytime iterative system to provide paths
between multiple objectives and to determine the visiting order of destinations;
moreover, the system should be informable meaning it can accommodate prior knowledge
of intermediate nodes, if available. The proposed system consists of two components:
\textbf{1)} an anytime informable multi-objective and multi-directional RRT$^*$
(\imomtN) algorithm to form a connected weighted-undirected graph, and \textbf{2)} a
relaxed TSP solver that consists of an enhanced version of the cheapest insertion
algorithm \cite{rosenkrantz1977analysis} and a genetic algorithm
\cite{potvin1996genetic, moon2002efficient, braun1990solving, ahmed2010genetic},
which together we call ECI-Gen. The proposed system is evaluated on large-complex
graphs built for real-world driving applications, such as the OpenStreetMap of
Chicago containing $866,089$ nodes and $1,038,414$ edges that is shown in
Fig.~\ref{fig:FirstImg}.


\begin{figure}[t]
    \centering
    \includegraphics[width=1\columnwidth]{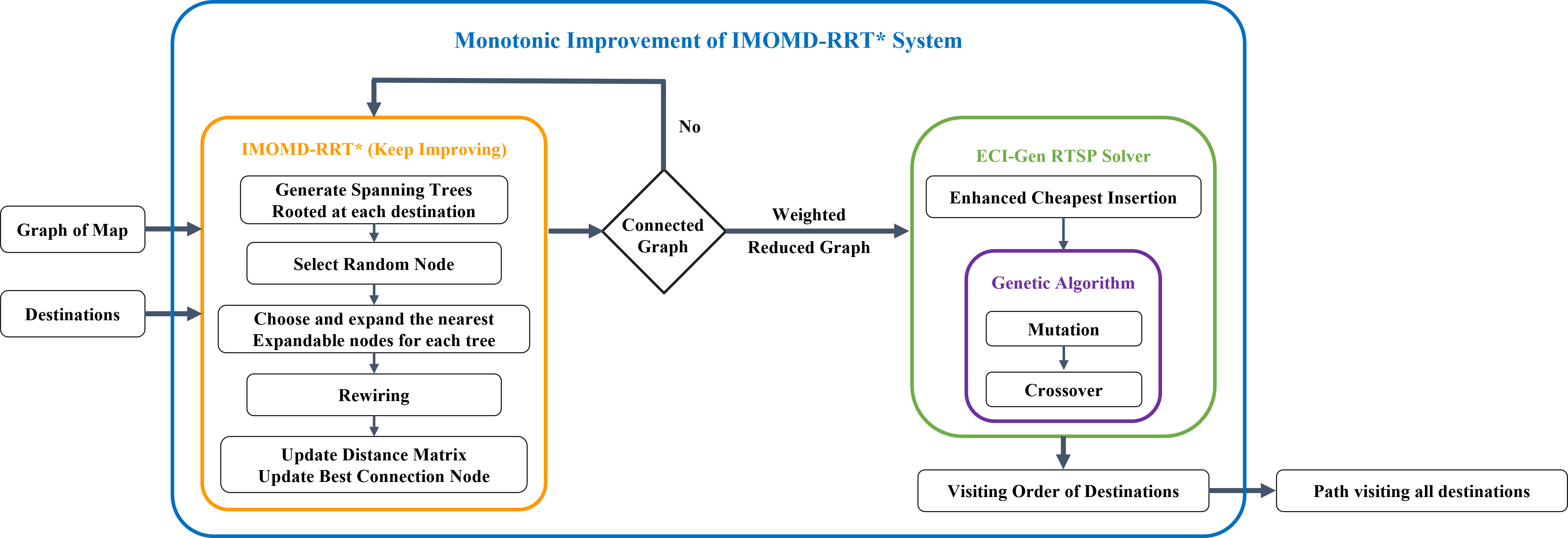}
    \caption{Illustration of the proposed \imomd system. It consists of an anytime informable multi-objective and
multi-directional RRT$^*$ (\imomtN) algorithm to construct a connected weighted-undirected
graph, and a polynomial-time solver to solve the relaxed TSP. The solver 
consists of an enhanced version of the cheapest insertion algorithm \cite{rosenkrantz1977analysis} and a genetic algorithm \cite{potvin1996genetic, moon2002efficient, braun1990solving}, called ECI-Gen solver. The full system (the blue box) will continue to run to further improve the solution over time.}
    \label{fig:SystemDiagram}
\end{figure}



The overall system is shown in Fig.~\ref{fig:SystemDiagram} includes the following
contributions: 
\vspace{-2mm}
\begin{enumerate}
    \item An anytime informable multi-objective and multi-directional RRT$^*$
        (\imomtN) system that functions on large-complex graphs. The anytime features means that the system quickly constructs a path on a large-scale undirected weighted
        graph that meets the existence constraint (the solution path must
        passes by all the objectives at least once), and the order constraint (with
        fixed starting point and end point). Therefore, the resulting weighted reduced graph containing the
        objectives, source, and target is connected.
    \item The problem of determining the visiting order of destinations of the connected graph
        is a relaxed TSP (R-TSP) with fixed start and end nodes, though intermediate nodes can be (or sometimes must be) visited more than once.  We
        introduce the \tspsolverN, which is based on an enhancement of the cheapest
        insertion algorithm \cite{rosenkrantz1977analysis} and a genetic
        algorithm \cite{potvin1996genetic, moon2002efficient, braun1990solving,
        ahmed2010genetic} to solve the R-TSP in polynomial time.
    \item We show that prior knowledge (such as a reference path) for robotics
        inspection of a pipe or factory can be readily and inherently integrated
        into the \imomtN. In addition, providing the prior knowledge to the
        planner can help navigate through challenging topology.
    \comment{
    \item \bdd{We benchmark the \tspsolver on various orders of graphs with the brute force algorithm\cite{cormen2022introduction} that permutes all possible solutions and an open source TSP solver\cite{concorde2005tsp}. The results shows that the proposed solver is able to reach optimal solutions on average \bdd{99}\% and \bdd{85}\% in the testing datasets of complete graphs and incomplete graphs, respectively.}
    }
    \item \conference{We evaluate the system comprised of \imomd and the \tspsolver on
        large-scale graphs built for real world driving applications, where the large number of intermediate destinations precludes solving the ordering by brute force. We show
        that the proposed IMOMD-RRT$^\ast$ outperforms bi-directional A$^*$ \cite{BiAstar} and ANA$^*$ \cite{ANAstar}
        in terms of speed and memory usage in large and complex graphs. We also demonstrate by providing a reference path, the \imomt escapes from bug traps (e.g., single entry neighborhoods) in complex 
        graphs.}
        \comment{We extensively evaluate the system comprised of \imomd and \tspsolver on large-scale graphs built for real world applications . In particular, we show that the proposed \imomt outperforms \bdd{Dijkstra}\cite{dijkstra}, bi-directional A$^*$ \cite{BiAstar}, and ANA$^*$ \cite{ANAstar} in terms of speed and the number of explored nodes in large and complex graphs. In addition, we also demonstrate that due to the random sampling nature of \rrtN, the \imomt liberates from bug traps (single entry neighborhood) in complex graphs.}
    \item We open-source the multi-threaded C++ implementation of the system at\\
\href{https://github.com/UMich-BipedLab/IMOMD-RRTStar}{https://github.com/UMich-BipedLab/IMOMD-RRTStar}.
\end{enumerate}

\conference{
The remainder of this paper is organized as follows. Section~\ref{sec:RelatedWork}
summarizes the related work. Section~\ref{sec:IMOMDRRT} explains the
proposed anytime informable multi-objective and multi-directional RRT$^*$ to
construct a simple connected graph. The \tspsolver to determine the ordering of destinations in the connected graph is discussed in Sec.~\ref{sec:WIBSFRTSP}. Experimental evaluations of
the proposed system on large-complex graphs are presented in
Sec.~\ref{sec:Experiments}. Finally, Sec.~\ref{sec:Conclusion} concludes the paper
and provides suggestions for future work.
}

\section{Related Work}
\label{sec:RelatedWork}
Path planning is an essential component of robot autonomy. In this section, we review several types of path planning algorithms and techniques to improve
their efficiency. Furthermore, we compare the proposed system with existing literature on
car-pooling/ride-sharing and the traveling salesman problem.

\subsection{Common Path Planners}
\label{sec:CommonPathPlanners}
Path planners are algorithms to find the shortest path from a single source to a
single target. Graph-based and sampling-based algorithms are the two prominent categories.


\conference{
Graph-based algorithms \cite{AStar, BiAstar, dijkstra, Rstar, ARAstar, ANAstar, JPS} 
such as Dijkstra\cite{dijkstra} and
\Astar\cite{AStar} discretize a continuous space to an
undirected graph composed of nodes and weighted edges. They are popular for their
efficiency on low-dimensional configuration spaces and small graphs.
There are many techniques\cite{BiAstar, Rstar, ARAstar, ANAstar, JPS} to improve
their computation efficiency on large graphs. Inflating the heuristic value makes the
\Astar algorithm likely to expand the nodes that are close to the goal and results in
sacrificing the quality of solution. Anytime Repairing \Astar (ARA$^*$)
\cite{ARAstar} utilizes weighted \Astar and keeps decreasing the weight parameter
at each iteration, and therefore leads to a better solution. Anytime Non-parametric
\Astar (ANA$^*$) \cite{ANAstar} is as efficient as ARA$^*$ and spends less time between solution improvements. R$^*$
\cite{Rstar} is a randomized version of \Astar to improve performance. Algorithms
such as Jumping Point Search \cite{JPS} to improve exploration efficiency only work
for grid maps. However, graph-based algorithms
inherently suffer from bug traps, whereas sampling-based methods can overcome bug traps more easily via informed sampling; see Sec.~\ref{sec:Experiments} for a detailed discussion.
}

\comment{
Graph-based algorithms \cite{AStar, BiAstar, dijkstra, Rstar, ARAstar, ANAstar, JPS,
LifelongAstar, Dstar, Dstarlite} such as Dijkstra\cite{dijkstra} and
\Astar\cite{AStar} discretize a continuous space and represent the space as an
undirected graph composed of nodes and weighted edges. They are popular for their
optimality and efficiency on low dimensional configuration spaces and small graphs.
There are many techniques\cite{BiAstar, Rstar, ARAstar, ANAstar, JPS} to improve
their computation efficiency on large graphs. Inflating the heuristic value makes the
\Astar algorithm likely to expand the nodes that close to the goal and results in
sacrificing the quality of solution. Anytime Repairing \Astar (ARA$^*$)
\cite{ARAstar} utilizes weighted \Astar and keeps decreasing the weight parameter
every iteration, and therefore leads to a better solution. Anytime Non-parametric
\Astar (ANA$^*$) \cite{ANAstar} is a non-parameter version of ARA$^*$. R$^*$
\cite{Rstar} is a randomized version of \Astar to improve performance. Algorithms
such as Jumping Point Search \cite{JPS} to improve exploration efficiency only work
for grid maps with or without diagonal action. However, graph-based algorithms
inherently suffer from bug traps, whereas sampling-based methods do not; see Sec.~\ref{sec:Informable} and Sec.~\ref{sec:BugTraps} for detailed discussion. 
}

Sampling-based algorithms such as rapidly-exploring random trees (RRT)
\cite{VanillaRRT} stand out for their low complexity and high efficiency in exploring
higher-dimensional, continuous configuration spaces. Its asymptotically optimal
version -- \rrt \cite{RRTStar, RRTStarICRA, huang2021efficient} -- has also gained much
attention and has contributed to the spread of the RRT family. More recently,
sampling-based algorithms on discrete spaces such as RRLT and d-RRT$^*$ have been applied to multi-robot
motion planning\cite{morgan2004sampling, branicky2003rrts, dRRT, dRRTstar}. 
We seek to leverage \rrt to construct a simple connected graph that contains multiple destinations from a large-complex
map, as well as to accommodate prior knowledge of a reference path. 


\subsection{Car-Pooling and ride-sharing}
Problems such as car-pooling, ride-sharing, food delivery, or combining public
transportation and car-pooling handle different types of constraints such as maximum
seats, time window, battery charge, number of served requests along with
multiple destinations \cite{ma2018path, huang2018multimodal, al2019deeppool,
huang2018ant, duan2018optimizing, tamannaei2019carpooling,
hulagu2020electric,suman2019improvement, lyu2019cb, simoni2020optimization,
naccache2018multi, nazari2018reinforcement, lu2019hybrid}. These problems are usually
solved by Genetic Algorithms \cite{ma2018path}, Ant Colony Optimization
\cite{huang2018ant, lu2019hybrid}, Dynamic Programming \cite{lyu2019cb,
simoni2020optimization} or reinforcement learning \cite{al2019deeppool,
nazari2018reinforcement}. These methods assume, however, that weighted paths between destinations in the graph are already known\cite{ma2018path, huang2018ant, lu2019hybrid, simoni2020optimization, al2019deeppool, duan2018optimizing, suman2019improvement}, while in practice, the connecting paths and their weights are unknown and must be constructed.

\begin{figure}[t]%
    \centering
    \subfloat[]{%
        \label{subfig:Acyclic}%
    \includegraphics[trim=0 0 0 0,clip,width=0.18\columnwidth]{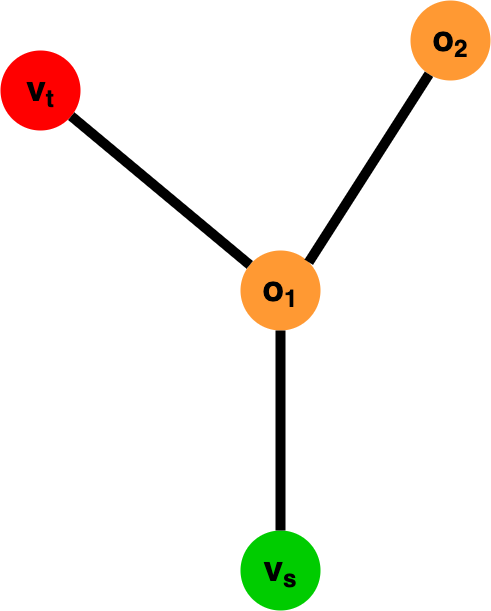}}\hspace{2cm}
        \subfloat[]{%
        \label{subfig:BadHamiltonian}%
    \includegraphics[trim=0 0 0 0,clip,width=0.18\columnwidth]{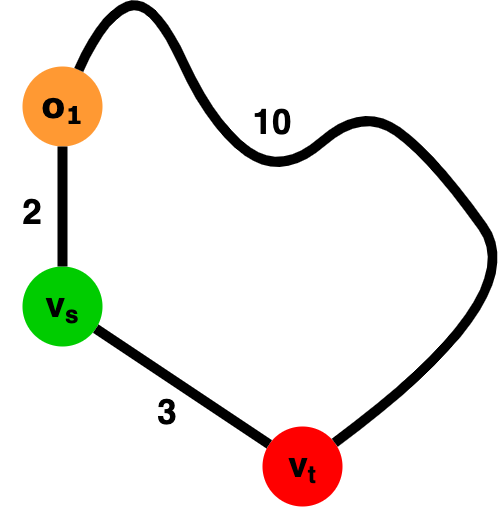}}%
    \caption[]{\subref{subfig:Acyclic} shows a \textit{simple
            connected graph} where the objectives 
    have to be visited twice to visit all destinations.
\subref{subfig:BadHamiltonian} shows the case where revisiting the source $\vsub{s}$
allows a shorter path when triangular inequality does not hold.
}%
\label{fig:Hamiltonian}%
    \squeezeup
\end{figure}

\subsection{Multi-objective Path Planners and Traveling Salesman Problem}
\label{sec:MO-TSP}
Determining the travel order of nodes in an undirected graph with known edge weights
is referred to as a Traveling Salesman Problem (TSP). The TSP is a classic NP-hard
problem about finding a shortest possible cycle that visits every node exactly once
and returns to the start node \cite{applegate2011traveling, junger1995traveling}. 
Another variant of TSP, called the shortest Hamiltonian path
problem\cite{junger1995traveling}, is to find the shortest path that visits all nodes
exactly once between a fixed starting node ($v_s$) and a fixed terminating node
($v_t$). The problem can be solved as a standard TSP problem by assigning
sufficiently large negative cost to the edge between $v_s$ and $v_t$
\cite{junger1995traveling, applegate2011traveling}. 

We are inspired by the work of \cite{vonasek2019space, janovs2021multi}, where the
authors propose to solve the problem through a single-phase algorithm in
simulated continuous configuration spaces. They first leverage multiple random trees
to solve the multi-goal path planning problem and then solve the TSP via an
open-source solver. In particular, they assume that the paths between nodes in the
continuous spaces can be constructed in single pass and that the resulting graph can
always form a cycle in their simulation environment. Thus, the problem can then be
solved by a traditional TSP solver. In practice, however, the graph might not form a
cycle (e.g., an acyclic graph or a forest) as shown in Fig.~\ref{subfig:Acyclic}, and
even if the graph is cyclic or there exists a Hamiltonian path, there is no guarantee
the path is the shortest. In Fig.~\ref{subfig:BadHamiltonian}, the Hamiltonian path
simply is $v_s\rightarrow o_1\rightarrow v_t$, and the traversed distance is $12$.
However, another shorter path exists if we are allowed to traverse a node ($v_s$ in
this case) more than once: $v_s\rightarrow o_1\rightarrow v_s\rightarrow v_t$ and the
distance is $7$. We therefore propose a polynomial-time solver for this relaxed TSP
problem; see Sec.~\ref{sec:WIBSFRTSP} for further discussion.




\section{Informable Multi-objecticve and Multi-directional RRT$^\ast$}
\label{sec:IMOMDRRT}
This section introduces an anytime informable multi-objective and multi-directional
Rapidly-exploring Random Tree$^*$ (\imomdN) algorithm as a real-time
means to quickly construct from a large-scale map a weighted undirected graph that
meets the existence constraint (the solution trajectory must pass by all the
objectives at least once), and the order constraint (with fixed starting point and
end point). In other words, the \imomd forms a simple\footnote{A simple graph, also called a strict graph, is an unweighted and undirected graph that contains no graph loops or multiple edges between two nodes\cite{bondy1976graph}.} connected graph containing the objectives, source, and target.



\subsection{Standard \rrt Algorithm}
The original \rrt\cite{RRTStar} is a sampling-based planner with
guaranteed asymptotic optimality in continuous configuration spaces. In general, \rrt
grows a tree where nodes are connected by edges of linear path segments. Additionally, \rrt considers nearby nodes of a newly extended node when choosing the best parent node and when rewiring the graph to find shorter paths for asymptotic optimality.



\subsection{Multi-objective and Multi-directional \rrt on Graphs}
\label{sec:RRTOnGraphs}
\begin{figure}[t]%
    \centering
    \subfloat[\label{subfig:expandable}]{%
    \includegraphics[trim=0 0 0 0,clip,width=0.30\columnwidth]{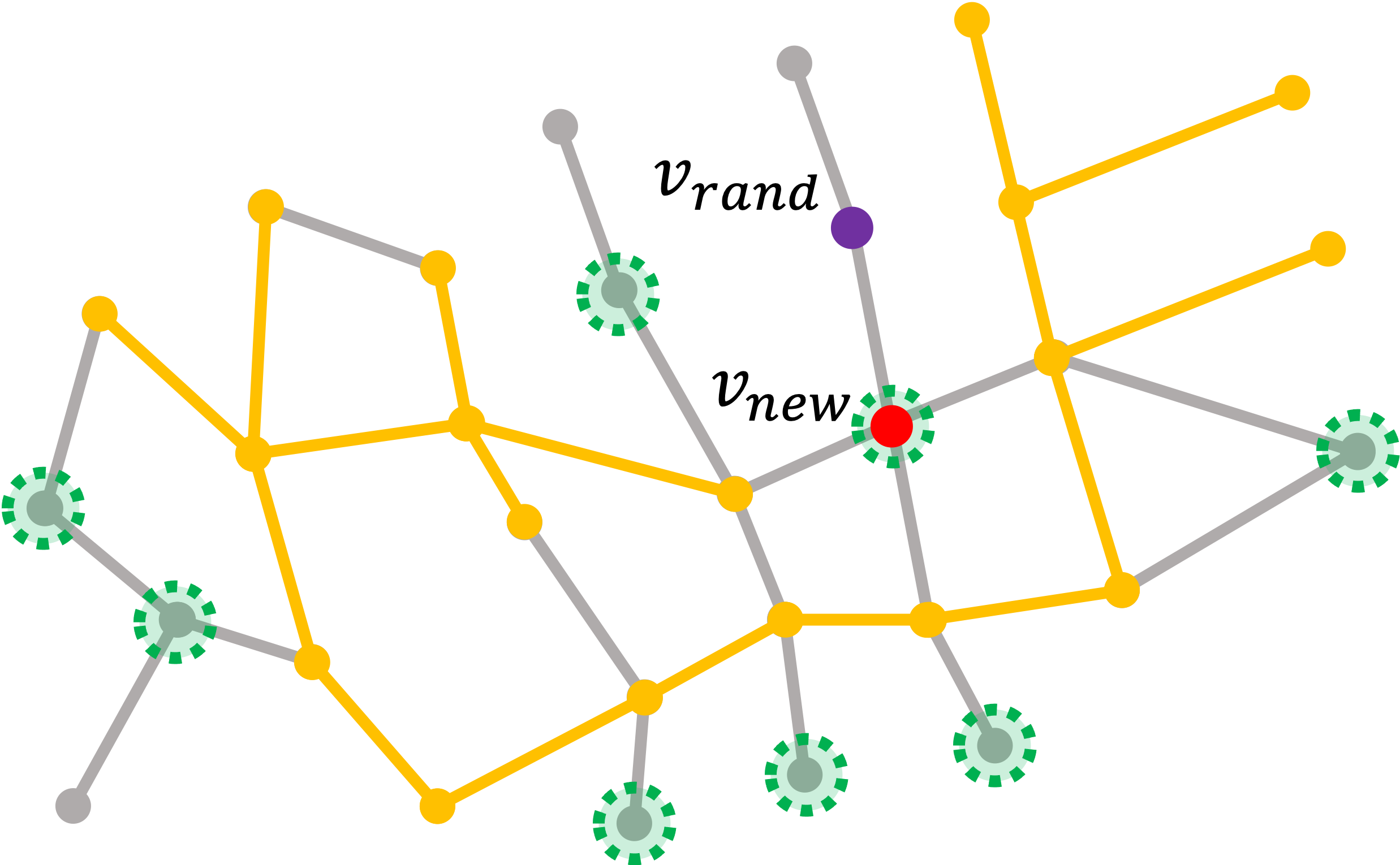}}~
    \subfloat[\label{subfig:rewiring}]{%
    \includegraphics[trim=0 0 0 0,clip,width=0.30\columnwidth]{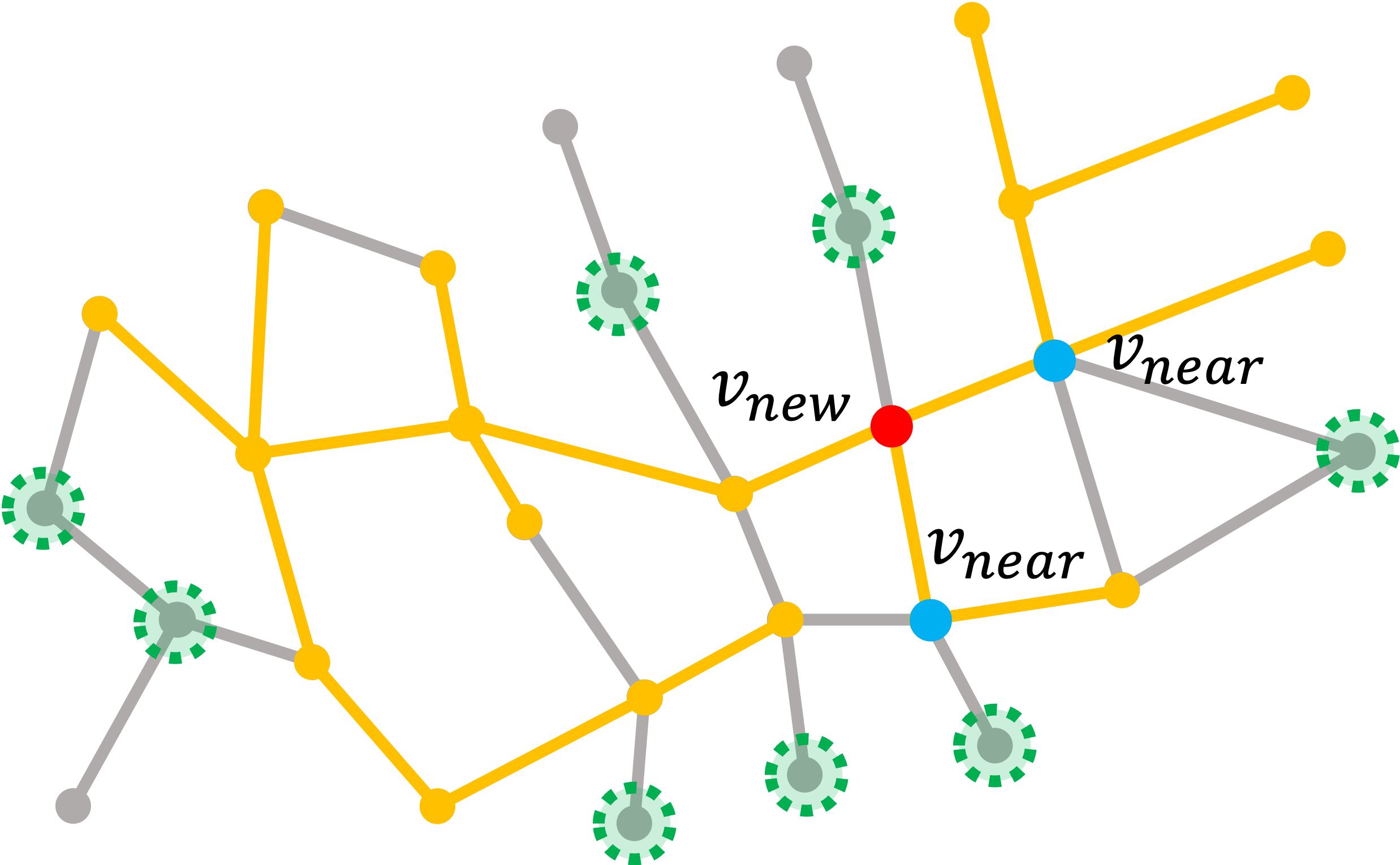}}
    \caption[]{{Illustration of tree expansion and connection nodes of a tree. The unexplored paths in the graph and the spanning tree are represented by the gray and yellow lines, respectively. The green-dashed circles are the expandable nodes. \subref{subfig:expandable} shows the tree $\Tcal_i$ extends to the $v_\text{rand}$ by growing a node $v_\text{new}$ to the closest expandable node.  \subref{subfig:rewiring} shows the updated set $\Xcal_i$ of expandable nodes and the tree is rewired around the $v_\text{new}$. The rewired nodes $v_\text{near}$ are represented as the blue dots.}}%
    \label{fig:expandable}%
    \squeezeup
\end{figure}

\begin{figure}[t]%
    \centering
    \subfloat[\label{subfig:connection}]{%
    \includegraphics[trim=0 0 0 0,clip,width=0.30\columnwidth]{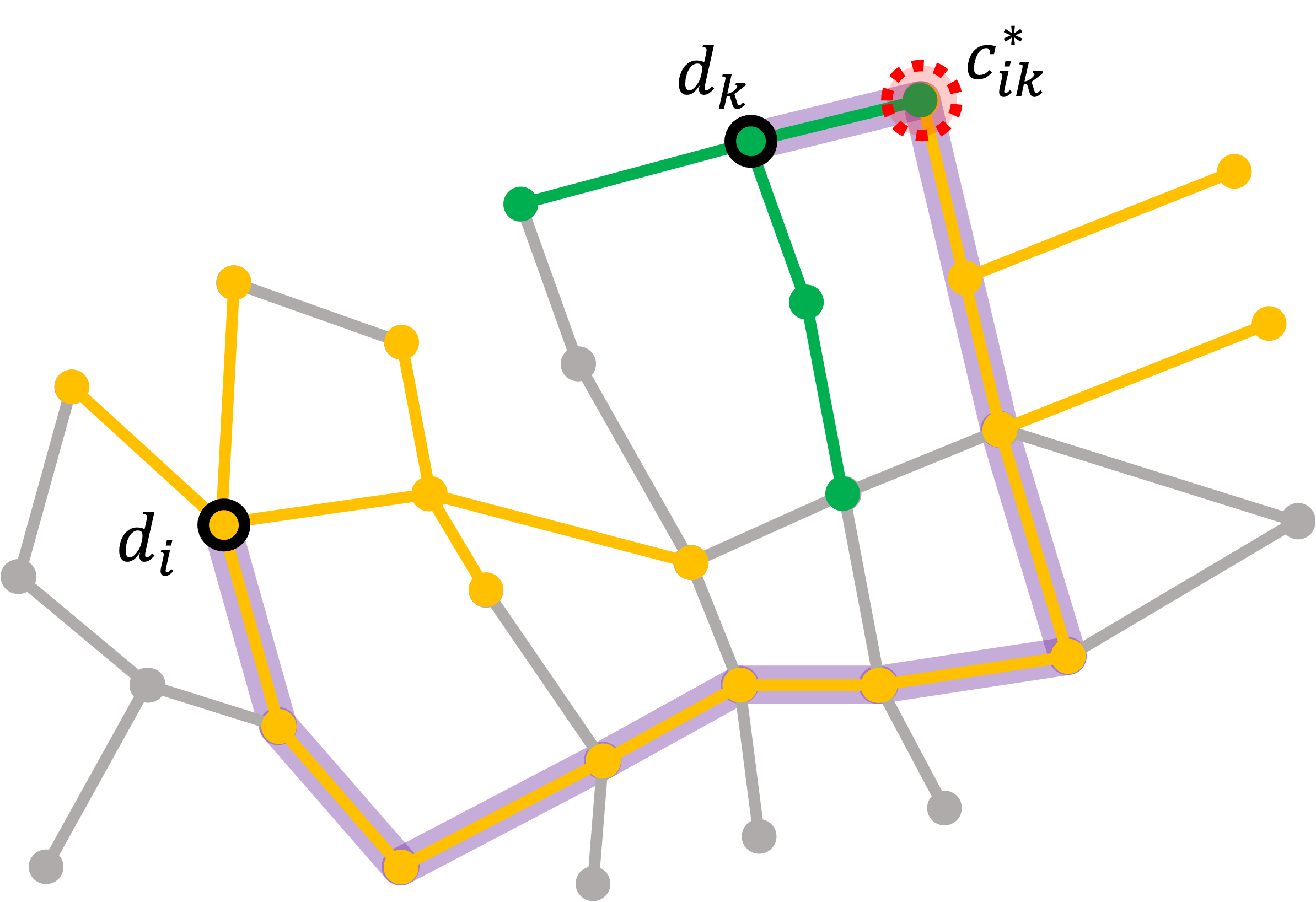}}~
    \subfloat[\label{subfig:update_connection}]{%
    \includegraphics[trim=0 0 0 0,clip,width=0.30\columnwidth]{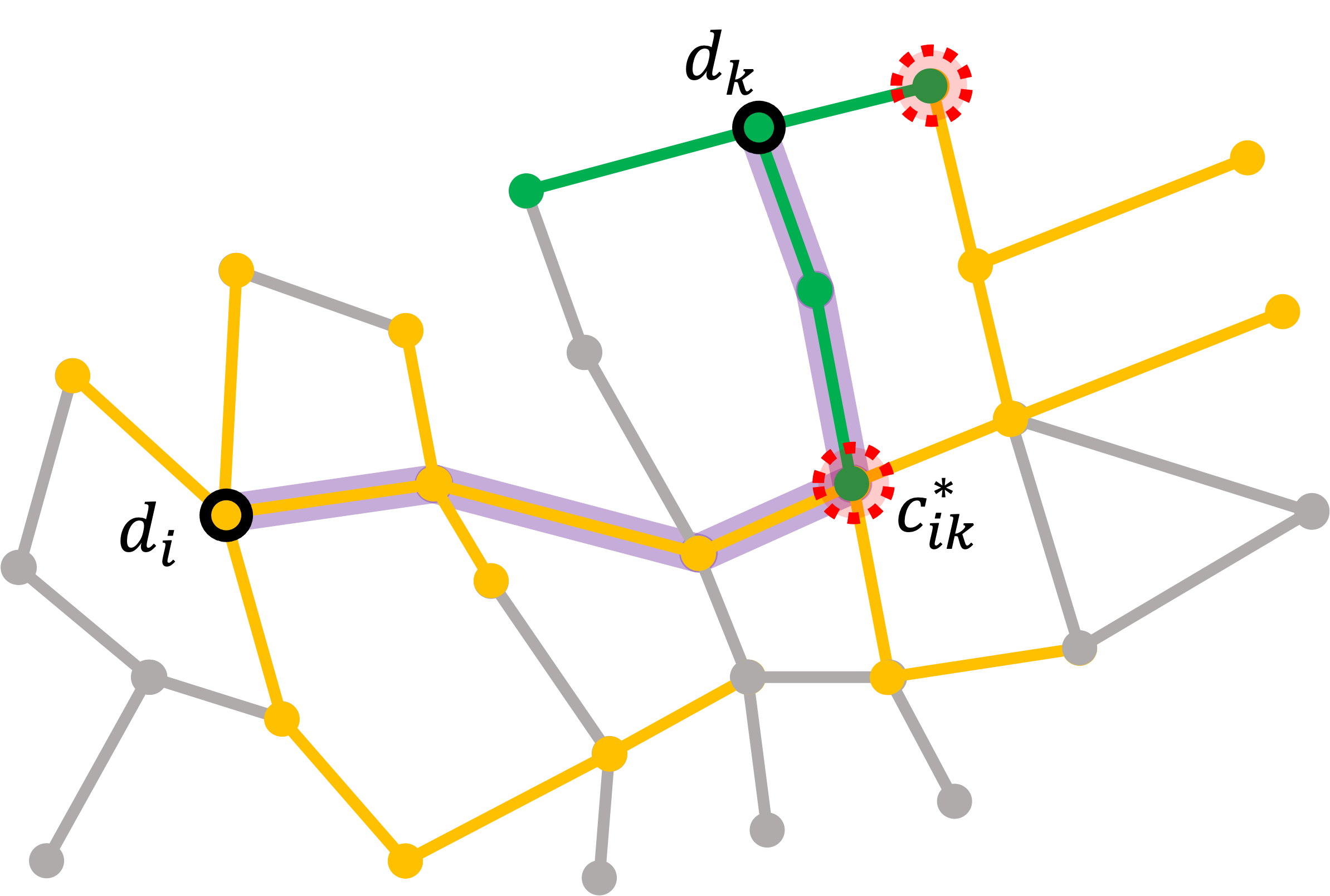}}~
    \caption[]{{Illustration of better connection nodes resulting in a better path. 
    Two trees rooted at $d_i$ and $d_k$ are marked in yellow and green, respectively. The highlighted purple line shows the shortest path between $d_i$ and $d_k$. The newly extended node $v_\text{new}$ (dashed-red circles) is added to a set of connection nodes $\Ccal_{ik}$. The element of the distance matrix, $A_{ik}$, and the connection node $c_{ik}^{*}$ that generates the shortest path between the destination $d_i$ and $d_k$ are updated as a shorter path is found.}}%
    \label{fig:UpdateConnection}%
    \squeezeup
\end{figure}

In this paper, we use \textit{map} to refer to the input graph, which might contain
millions of nodes, and use \textit{graph} to refer to the graph composed of only the
destinations including the source and target node.  The proposed \imomd differs from
the original \rrt in six aspects when growing a tree. First, the sampling is performed by picking a
random $v_\text{rand}$ in the map, and not from an underlying continuous space.
The goal bias is not only applied to the target but also the source and all the
objectives. Second, a steering function directly finds the closest expandable node as
$v_\text{new}$ to the random node $v_\text{rand}$, without finding the nearest node
in the tree, as shown in Fig. \ref{subfig:expandable}. Note that instead
of directly sampling from the set of expandable nodes, sampling from the map
ameliorates the bias of sampling on the explored area. Third, the parent node is
chosen from the nodes connected with the new node $v_\text{new}$, called the neighbor
nodes. Among the neighbor nodes, the node that yields the lowest path cost from
the root becomes the new node's parent. Fourth, the jumping point search
algorithm~\cite{JPS} is also leveraged to speed up tree exploration. Fifth, the
\imomd rewires the neighborhood nodes to minimize the accumulated cost from the root
of a tree to $v_\text{new}$, as shown in Fig. \ref{subfig:rewiring}. {Lastly, if
$v_\text{new}$ belongs to more than one tree, this node is considered a connection
node, which connects the path between destinations, as shown in Fig.
\ref{fig:UpdateConnection}}.

Our proposed graph-based \rrt modification is summarized below with notation that
generally follows graph theory~\cite{bondy1976graph}. A graph $\Gcal$ is an ordered
triple $(\Vcal(\Gcal), \Ecal(\Gcal), \Phi_\Gcal)$, where
$\Vcal(\Gcal) = \{v\in\zeta\}$ 
is a set of nodes in the robot state space $\zeta$, $\Ecal(\Gcal)$ is a set of edges
(disjoint from $\Vcal(\Gcal)$), and an indication function $\Phi_\Gcal$ that
associates each edge of $\Gcal$ with an unordered pair (not necessarily distinct) of
nodes of $\Gcal$.

Given a set of destinations $\Dcal = \{d_i|d_i\in\{v_s, v_t\} \cup \Ocal \}_{i=1}^{m+2}$, 
where $v_s\in\Vcal(\Gcal)$ is the source node, $v_t\in\Vcal(\Gcal)$ is the target node,
and $\Ocal\subseteq\Vcal(\Gcal)$ is the set of $m$ objectives, the \imomd solves
the multi-objective planning problem by growing a tree $\Tcal_i = (V, E)$, where
$V\subseteq\Vcal(\Gcal)$ is a set of nodes connected by edges $E\subseteq\Ecal(\Gcal)$,
at each of the destinations $d_i\in\Dcal$. Thus, it leads to a family of trees $\Tcal
= \{\Tcal_1, \cdots, \Tcal_m, \Tcal_{m+1}, \Tcal_{m+2} \}$. 


%
%
%
%
%
%
%
%
%
%
%

\conference{
The proposed \imomd explores the graph $\Gcal$ by random
sampling from $\Vcal(\Gcal)$ and extending nodes to grow each tree. We explain a few important functions of IMOMD-RRT$^\ast$ below.}

\subsubsection{Tree Expansion}
Let ${\Ncal(v)}$ be the set of nodes directly connected with a node, i.e., the neighborhood of a node $v$~\cite{bondy1976graph}. 
{A node is expandable if there exists at least one unvisited node connected and at least one node of the tree connected, shown as the dashed-green circles in Fig~\ref{subfig:expandable}}. Let $\Xcal_i$ be the set of expandable 
nodes of the tree $\Tcal_i$. A random node $v_\text{rand}$ is sampled from
the nodes of the graph $\Vcal(\Gcal)$. Next, find the nearest node $v_\text{new}$
in the set of expandable nodes $\Xcal_i$:
\begin{equation}
\label{eq:NearestNode}
    v_\text{new} = \underset{v \in  \Xcal_i }{\argmin} \, \mathtt{Dist}(v, v_\text{rand}),
\end{equation}
where $\mathtt{Dist(\cdot, \cdot)}$ is the distance between two states.
Next, the jumping point search algorithm\cite{JPS} is utilized to speed up the tree expansion. If the current $v_\text{new}$ has only one neighbor that is not already in the tree, $v_\text{new}$ is added to the tree and that one neighbor is selected as the new $v_\text{new}$. This process continues until $v_\text{new}$ has at least two neighbors that are not in tree, or it reaches $v_\text{rand}$.

\subsubsection{Parent Selection}
Let the set $\Ncal_i({v_\text{new})}$ be the neighborhood of the $v_\text{new}$ in
the tree $\Tcal_i$. The node $v_\text{near}$ in $\Ncal_i({v_\text{new})}$ that
results in the smallest cost-to-come, $\mathtt{Cost(\cdot,
\cdot)}$, is the parent of the
$v_\text{new}$ and is determined by: \begin{equation} v_\text{parent} =
\underset{v_\text{near} \in \Ncal_i({v_\text{new})}}{\argmin} \{
\mathtt{Cost}(\Tcal_i, v_\text{near}) + \mathtt{Dist}(v_\text{near}, v_\text{new})
\}. \end{equation} \conference{Next, all the unvisted nodes in
$\Ncal_i({v_\text{new})}$ are added to the set of expandable nodes $\Xcal_i$.}
\comment{Next, all the unvisted nodes in $\Qcal_i$ are added to the set of expandable
nodes $\Xcal_i$; see Line 3 in Algorithm \ref{alg:extend}.}



\subsubsection{Tree Rewiring}
\conference{After the parent node is chosen, the nearby nodes are rewired if a
shorter path reaching the node through the $v_\text{new}$ is found, as shown in
Fig. \ref{subfig:rewiring}. The rewiring step guarantees asymptotic optimality, as
with the classic algorithm.}

\subsubsection{Update of Tree Connection}
A node is a connection node if it belongs to more than one tree. Let $\Ccal_{ik}$ be the set of connection nodes between $\Tcal_i$ and $\Tcal_k$, and let $c_{ik}^*$ denote the node that connects $\Tcal_i$ and $\Tcal_k$ with the shortest distance
\begin{equation}
    c_{ik}^{*} = \underset{c \in \Ccal_{ik}}{\argmin}~\{ \mathtt{Cost}(\Tcal_i, c) + \mathtt{Cost}(\Tcal_k, c) \}.
\end{equation}
Let $A_{(m+2)\times (m+2)}$ be a distance matrix that represents pairwise distances between the destinations, where $m$ is the number of objectives. The element $A_{i,k}$ indicates the shortest path between destinations $d_i$ and $d_k$, as shown in Fig.~\ref{subfig:update_connection}. $A_{i,k}$ is computed as
\begin{equation}
     A_{i,k} = \mathtt{Cost}(\Tcal_i, c_{ik}^{*}) + \mathtt{Cost}(\Tcal_k, c_{ik}^{*}).
\end{equation}

\subsection{Discussion of Informability} 
\label{sec:Informable}
\begin{figure}[t]%
    \centering
    \subfloat[\label{subfig:real_destinations}]{%
    \includegraphics[trim=0 0 0 0,clip,width=0.42\columnwidth]{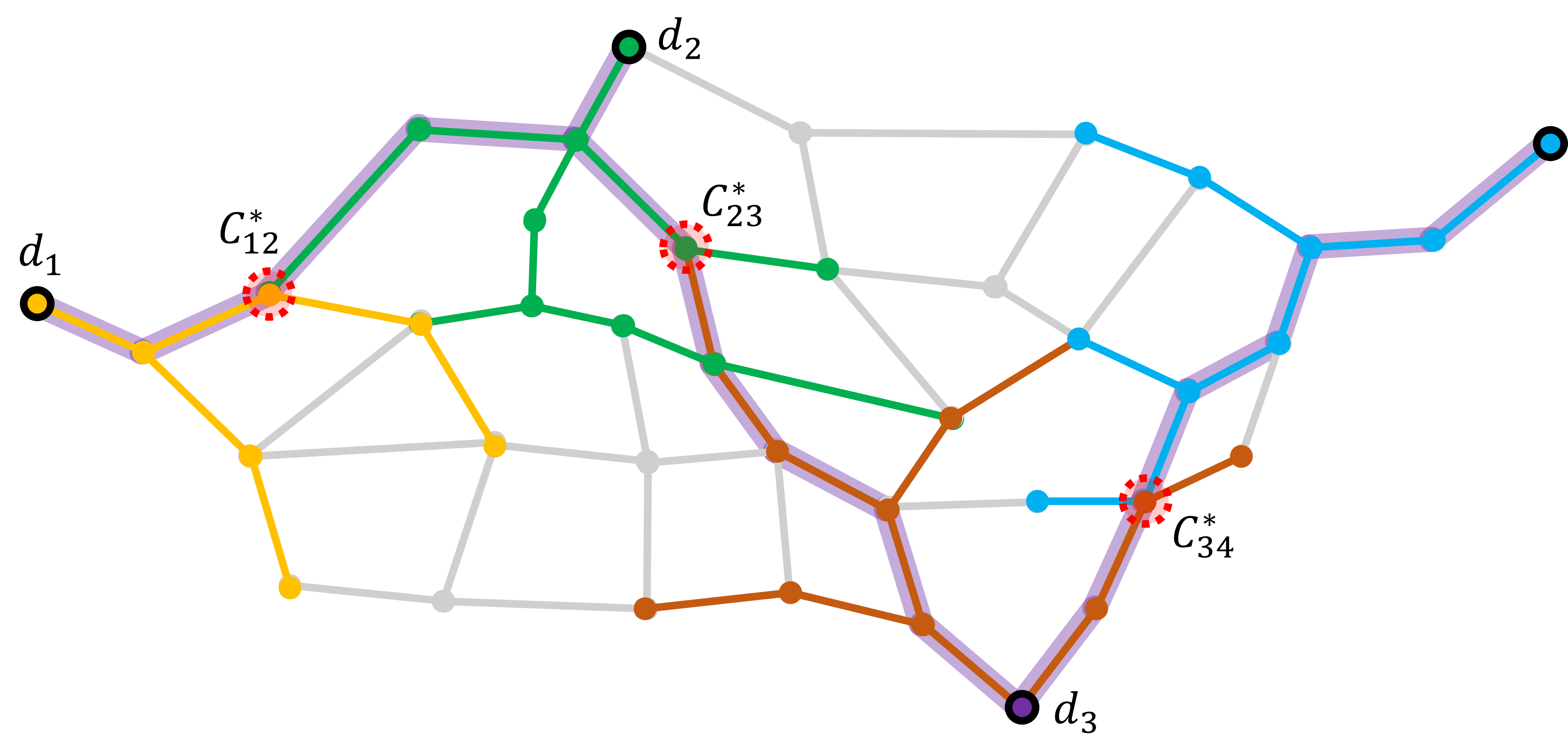}}
    \subfloat[\label{subfig:pseudo_destinations}]{%
    \includegraphics[trim=0 0 0 0,clip,width=0.42\columnwidth]{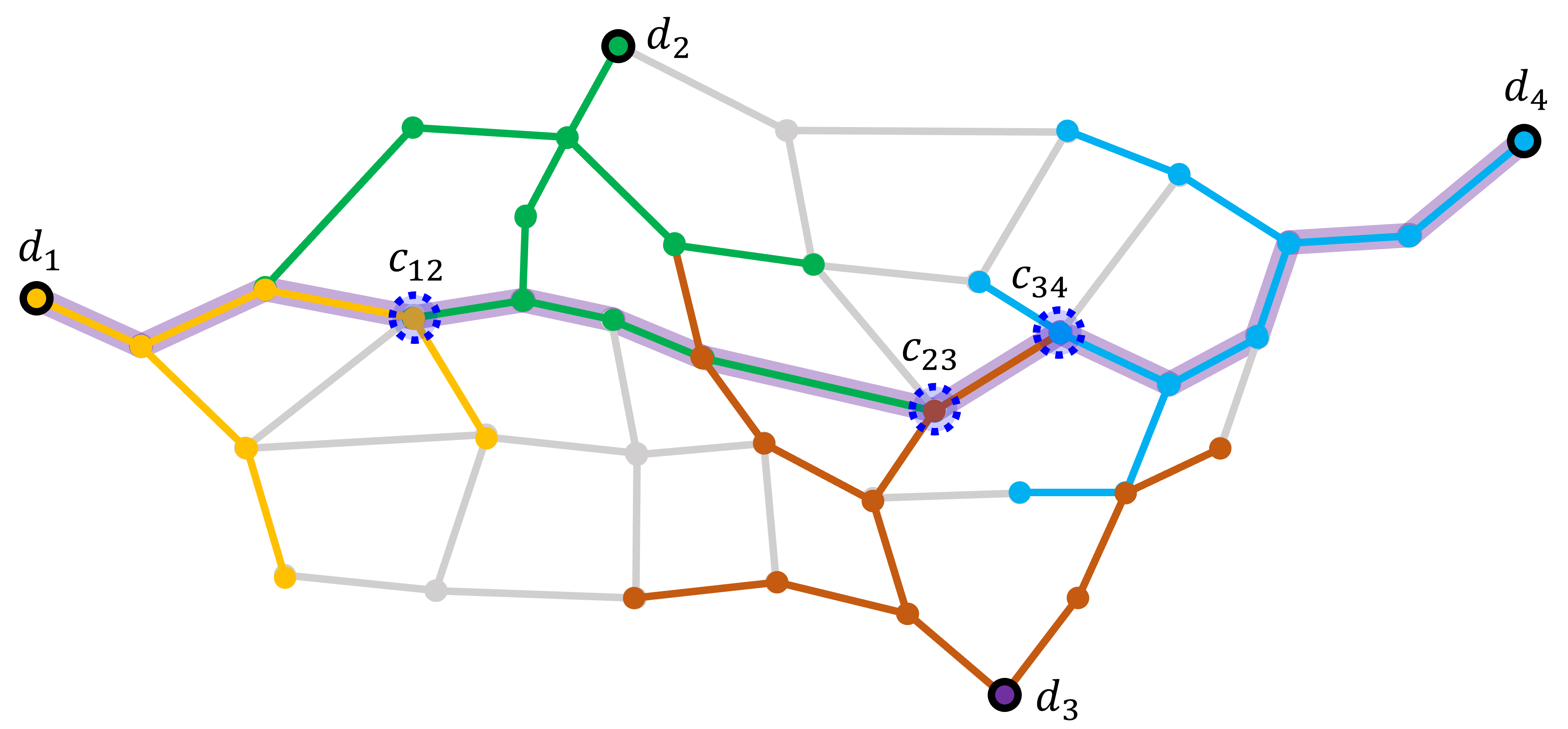}}
    \caption[]{{Illustration of the rewired path of pseudo-destinations. $d_1-d_4$ are the destinations and marked in different colors. The dashed-red circles are the connection nodes between two trees. The thick purple line is the final path. \subref{subfig:real_destinations} shows the resulting path as if $d_1-d_4$ are ``true'' destinations, and \subref{subfig:pseudo_destinations} is the resulting path as if $d_2-d_3$ are ``pseudo'' destinations. The real destinations have to be visited as shown in \subref{subfig:real_destinations}, whereas pseudo destinations are artificial destinations to help form a connected graph, and might no longer be visited after rewiring}}%
    \label{fig:informability}%
    \squeezeup
\end{figure}

As mentioned in Sec.~\ref{sec:Intro}, applications such as robotic inspection or vehicle routing might consider prior knowledge of the path, so that the robot can examine certain
equipment or area of interests or avoid certain areas in a factory. The prior
knowledge can be naturally provided as a number of ``pseudo destinations'' or samples
in the \imomdN. A pseudo destination is an artificial destination to help \imomd to
form a connected graph. However, unlike \textit{true} destinations that will always be visited, a pseudo destination might not be visited after rewiring, as shown in Fig. \ref{fig:informability}. Prior knowledge through pseudo
destinations can also be leveraged to traverse challenging topology, such as bug-traps; see
Sec.~\ref{sec:Experiments}.

\begin{remark}
    One can decide if the order of the pseudo destinations should be fixed or even
the pseudo destinations should be objectives (i.e., they will not be removed in the
rewiring process.). 
\end{remark}



\section{Enhanced Cheapest Insertion and Genetic Algorithm}
\label{sec:WIBSFRTSP} 
This section introduces a polynomial-time solver for the relaxed traveling salesman
problem (R-TSP).

\subsection{Relaxed Traveling Salesman Problem}
\label{sec:RelaxedTSP}
\conference{
The R-TSP differs from standard TSP\cite{applegate2011traveling, junger1995traveling,
punnen2007traveling} in two perspectives. First, nodes are allowed to be visited more
than once, as mentioned in Sec.~\ref{sec:MO-TSP}. Second, we have a source node where
we start and a target node where we end. Therefore, the R-TSP can also be considered
a relaxed Hamiltonian path problem\cite{junger1995traveling, punnen2007traveling}. We
propose the \tspsolverN, which consists of an enhanced version of the cheapest
insertion algorithm \cite{rosenkrantz1977analysis} and a genetic algorithm
\cite{potvin1996genetic, moon2002efficient, braun1990solving} to solve the R-TSP. The
complexity of the proposed solver is $O(N^3)$, where $N$ is the cardinality of the
destination set $\Dcal$.}


\comment{The to-be-solved R-TSP differs from standard TSP\cite{applegate2011traveling,
junger1995traveling, punnen2007traveling} in two perspectives. First,
nodes are allowed to visit more than once, as mentioned in Sec.~\ref{sec:MO-TSP}.
Second, we have a source node where we start and a target node where we end.
Therefore, the R-TSP can also be considered a relaxed Hamiltonian path
problem\cite{junger1995traveling, punnen2007traveling}. We propose
an \tspsolver (Algorithm~\ref{alg:aci_gen}), which consists of an Enhanced version of the cheapest
insertion \cite{rosenkrantz1977analysis} and the genetic algorithm \cite{potvin1996genetic, moon2002efficient, braun1990solving} to solve the R-TSP. The complexity of the proposed 
solver is $O(N^3)$, where $N$ is the cardinality of the destination set $\Dcal$.}

\subsection{Graph Definitions and Connectivity}
\conference{In graph theory \cite{west2001introduction, bondy1976graph}, a graph is simple or strict if it has no loops and no two edges join the same pair of nodes. In addition, a path is a sequence of nodes in the graph, where consecutive nodes in the sequence are adjacent, and no node appears more than once in the sequence. A graph is connected if and only if there is a path between each pair of destinations. Once all the destinations form a simple-connected graph, there exists at least one path $\pi$ that passes all destinations $\Dcal$. We can then consider the problem as an R-TSP (see Sec.~\ref{sec:RelaxedTSP}), where we
have a source and target node as well as several objectives to be visited. Therefore, we impose the graph connectivity and simplicity as sufficient conditions to solve the R-TSP. The disjoint-set data structure \cite{cormen2022introduction, galil1991data} is implemented to verify the connectivity of a graph.} 



\subsection{Enhanced Cheapest Insertion Algorithm}
\label{sec:WIBFS}
The regular cheapest insertion algorithm\cite{rosenkrantz1977analysis} provides an
efficient means to find a sub-optimal sequence that guarantees less than twice the optimal sequence cost. However, it does not handle the case where revisiting the same node
makes a shorter sequence. Therefore, we propose an enhanced version of the cheapest insertion
algorithm, which comprises of a set of actions:
\textbf{1)} in-sequence insertion, $\lambda_\text{in-sequence}$, which is the regular
cheapest insertion; \textbf{2)} in-place insertion, $\lambda_\text{in-place}$, to allow
the algorithm to revisit existing nodes; and \textbf{3)} swapping insertion,
$\lambda_\text{swapping}$, which is inspired by genetic algorithms. Finally,
sequence refinement is performed at the end of the algorithm.

\begin{figure}[t]%
    \centering
    \subfloat[]{%
    \label{fig:inplace}%
    \includegraphics[trim=0 0 0 0,clip,width=0.20\columnwidth]{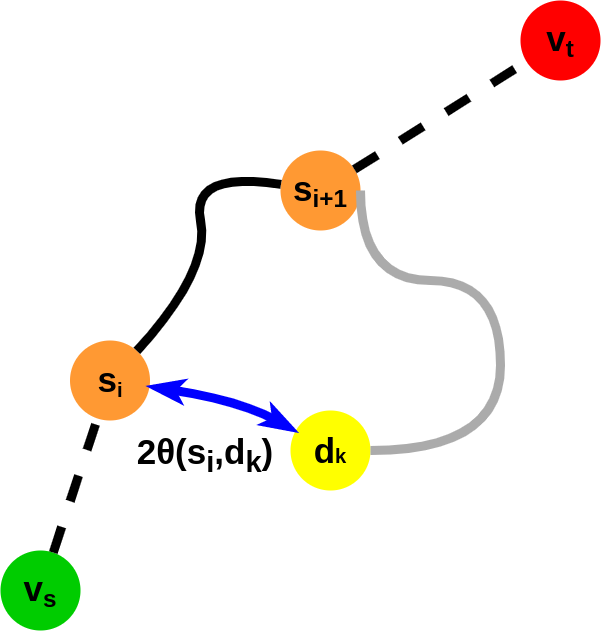}}~~~~~
    \subfloat[]{%
    \label{fig:insequence}%
    \includegraphics[trim=0 0 0 0,clip,width=0.20\columnwidth]{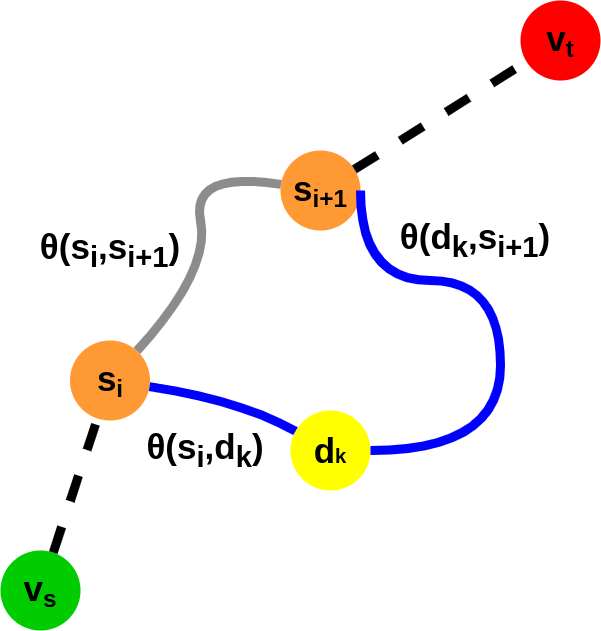}}%
    \caption[]{{Illustration of the insertion cost. \subref{fig:inplace} shows the in-place insertion and the resulting sequence contains duplicated $s_i$. \subref{fig:insequence} shows the in-sequence insertion without a duplicated element.}
    }%
    \label{fig:insertion}%
    \squeezeup
\end{figure}

Let the current sequence be $\Scal_{\text{current}} = \{ v_{s},s_{1}, \cdots,
s_{i},s_{i+1},\cdots, s_{n}, v_{t}\}$ to indicate the visiting order of destinations,
where $s_{\{\cdot\}}\in\Dcal$ and $(n+2)$ is the number of destinations in the sequence.
The travel cost $\theta(\cdot,\cdot)$ is the path distance between two destinations
provided by the \imomdN. The $\Scal_{\text{current}}$ is constructed by Dijkstra's algorithm on the graph.
\begin{remark}
    $\Scal_{\text{current}}$ possibly contains duplicated elements and $s_i$ is not
    necessary $d_i$. Therefore, the number of destinations in the sequence $n$ may be
    larger than the actual number of destinations $m$. Fig.~\ref{fig:inplace} shows
    the duplicated case with $\Scal_{\text{current}} = \{ v_s, s_1, \cdots, s_i,
    d_k, s_i, s_{i+1}, \cdots, s_n, v_t\}$, where $s_i$ is duplicated.
    Figure~\ref{fig:insequence} illustrates an unduplicated case where
    $\Scal_{\text{current}} = \{ v_s, s_1, \cdots,s_i, d_k, s_{i+1},$\\$\cdots, s_n, v_t\}$.
\end{remark}

Let $\Kcal$ denote the set of destinations to be inserted, and let the to-be-inserted
destination be $d_k$ and its ancestor be $s_i$, where $d_k\in\Kcal$ and
$s_i\in\Scal_\text{current}$. Given a current sequence $\Scal_{\text{current}}$, the location to insert $s^*$, and the action of insertion $\lambda^*$ are determined by
\begin{equation}
\label{eq:bestInsertionPlan}
    \lambda^*, s^* =
    \argmin_{\substack{\lambda_j\in\Lambda \\ 
                    s_i\in \Scal_\text{current}
              }
          }\lambda_j(s_i,d_k),
\end{equation}
where $\Lambda = \{\lambda_\text{in-sequence}, \lambda_\text{in-place}, \lambda_\text{swapping}\}$ is the set of the insertion actions.

\subsubsection{In-place Insertion $\lambda_\text{in-place}$}
This step detours from $s_{i}$ to $d_{k}$, and the resulting sequence is 
$\Scal_{\text{modified}} = \{ v_{s}, s_1 \cdots, s_{i}, d_{k}, s_{i}, s_{i+1},
\cdots, s_{n}, v_{t}\}$, as shown
in Fig.~\ref{fig:inplace}. The insertion distance is 
\begin{equation}
    \lambda_\text{in-place}(s_i,d_k)=2\theta(s_i,d_k).
\end{equation}

\subsubsection{In-sequence Insertion $\lambda_\text{in-sequence}$} 
    This step inserts $d_{k}$ between $s_{i}$ and $s_{i+1}$
    ($\forall s_i\in\{\Scal_\text{current}/(v_t)\}$), and the resulting sequence is
    $\Scal_{\text{modified}} = \{ v_{s}, s_{1}, \cdots, s_{i}, d_{k}, s_{i+1},
    \cdots, s_{n}, v_{t}\}$, as
    shown in Fig.~\ref{fig:insequence}. The insertion distance is
\begin{equation}
    \begin{aligned}
        \lambda_\text{in-sequence}(s_i,d_k) = \theta(s_i,d_k) + \theta(d_k, s_{i+1}) - \theta(s_i,s_{i+1}).
    \end{aligned}
\end{equation}


\subsubsection{Swapping Insertion $\lambda_\text{swapping}$}
    The swapping insertion
    changes the order of nodes right next to the newly inserted node. There are three cases in swapping insertion: swapping left, right, or both. For the case of swapping left, the modified sequence is
    $\Scal_\text{modified} = \{ v_{s}, s_1 \cdots, s_{i-2}, s_{i}, s_{i-1},
    d_{k}, s_{i+1}, \cdots, s_n, $\\$v_{t}\}$ by inserting $d_{k}$ between $s_{i}$ and 
    $s_{i+1}$ ($\forall s_i\in\{\Scal_\text{current}/(v_s, s_1)\}$), and then swapping $s_{i}$ and $s_{i-1}$. The insertion distance of swapping left is
    \begin{equation}
        \begin{aligned}
            \lambda_\text{swapping (left)}(s_i,d_k) &= \theta(d_k, s_{i+1}) - \theta(s_i,s_{i+1}) \\ 
                                                    &+ \theta(s_{i-1}, d_k) + \theta(s_{i-2}, s_i) \\
                                                    &- \theta(s_{i-2}, s_{i-1}).
        \end{aligned}
    \end{equation}
    
    \conference{The right swap is a similar operation except that it swaps $s_{i+1}$ and $s_{i+2}$ instead. Lastly, the case of swapping both does a left swap ($s_{i}$ and $s_{i+1}$) and then a right swap ($s_{i+1}$ and $s_{i+2}$).}

\subsubsection{Sequence Refinement}
\conference{
In-place insertion occurs when the graph is not cyclic or the triangular inequality does not
hold on the graph. The in-place insertion could generate redundant revisited nodes in the final result and lead to a longer sequence. We further refine the sequence by
{skipping revisited destination when the previous destination and the next destination are connected}. The refined sequence of destinations, $\Scal_\text{ECI}$, with cardinality $r\leq n$ is the input to the genetic algorithm.
}

\comment{
In-place insertion occurs when the graph is not cyclic or the triangular inequality does not
hold on the graph. The in-place insertion could generate redundant revisited nodes in final results and therefore leads to a longer path. We further refine the path by
{skipping revisited destination when the previous destination and the next destination are connected}, as shown in Algorithm~\ref{alg:refine_sequence}. The refined path $\Scal_\text{ACI}$ of the advanced cheapest insertion process is the input to the genetic algorithm; see the next section.
}

\subsection{Genetic Algorithm}
We further leverage a genetic algorithm \cite{potvin1996genetic, moon2002efficient,
braun1990solving} to refine the sequence from the enhanced cheapest insertion
algorithm. The genetic algorithm selects a parent\footnote{Note that the parent in
the genetic algorithm is a different concept from the parent node in the \rrt tree,
mentioned in Sec. \ref{sec:RRTOnGraphs}. The terminology is kept so that it follows
the literature consistently.} sequence and then generates a new offspring sequence from it by either a mutation or
crossover process.


\conference{
In brief, we first take the ordered sequence $\Scal_\text{ECI}$ from the enhanced
cheapest insertion as our first and only parent for the mutation process, which
produces multiple offspring. Only the offspring with a lower cost than the parent are
considered the parents for the crossover process.}

\comment{In brief, we first take the ordered sequence $\Scal_\text{ECI}$ from the advanced cheapest insertion as our first and only parent. Next, we apply the mutation process to the parent to produce multiple offspring, and only the offspring with a lower cost than the parent are kept. The remaining offspring are the parents for the crossover process, as shown in Fig.~\bh{ACI-Gen Algorithm}.}

\begin{figure}[t]%
    \centering
    \subfloat[Mutation]{
    \label{fig:Mutation}%
    \includegraphics[trim=0 0 0 0,clip,height=0.19\columnwidth]{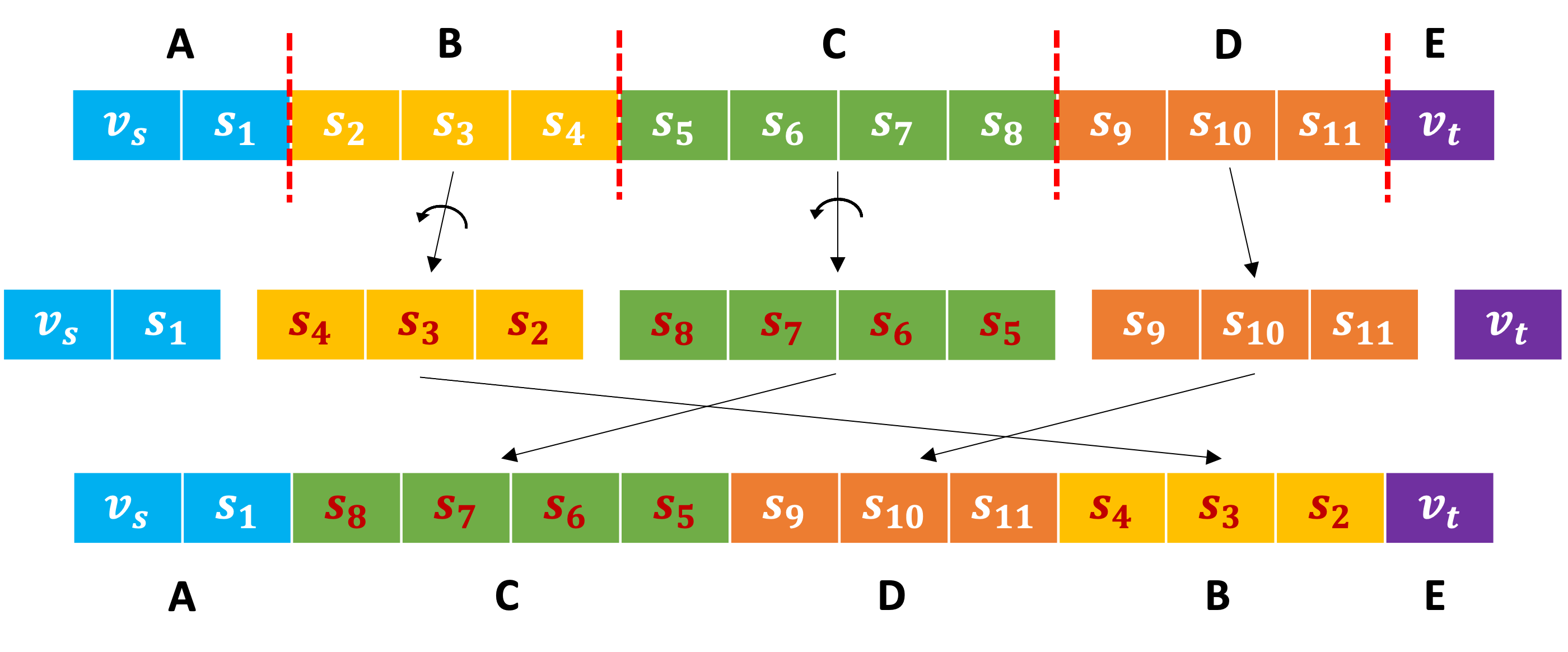}}~~
    \subfloat[Crossover]{
    \label{fig:Crossover}
    \includegraphics[trim=0 0 0 0,clip,height=0.19\columnwidth]{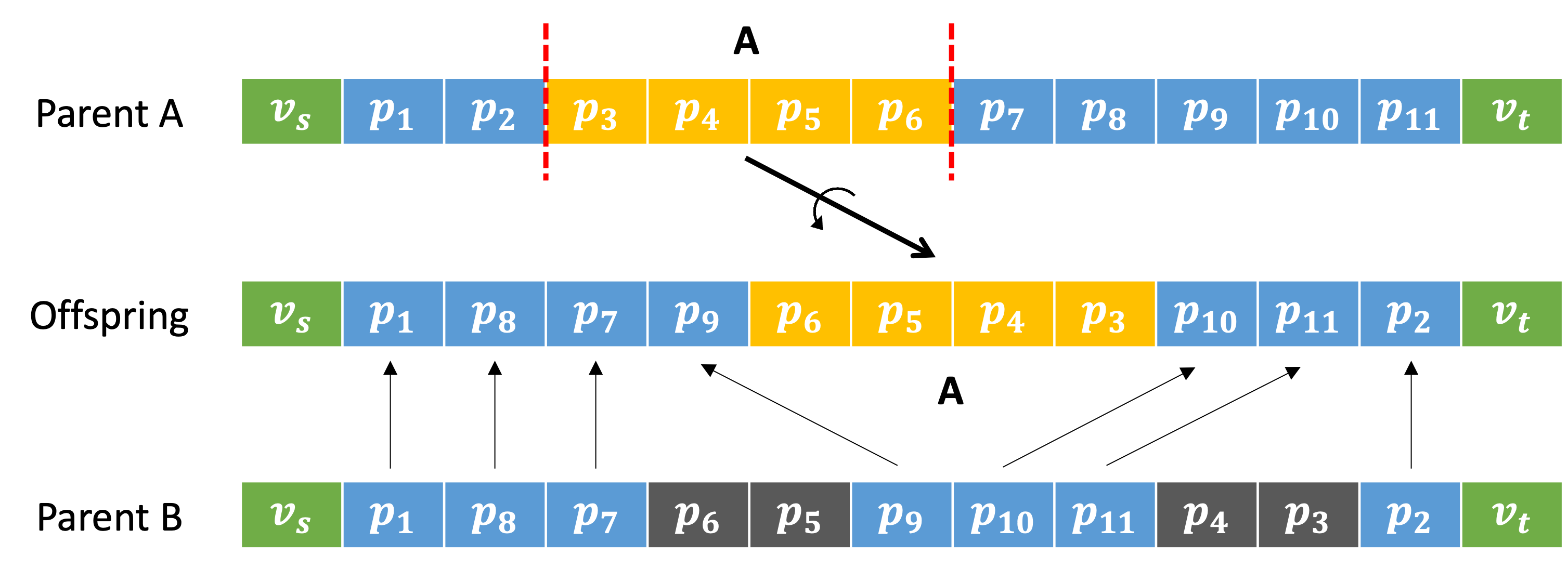}}%
    \caption[]{{Illustration of the mutation and crossover in the Genetic Algorithm. \subref{fig:Mutation} In mutation, the sequence is randomly cut into five
            segments and each segment is randomly reversed except Segment A and Segment E. In this case, Segment B and C are reversed and D does not. Finally, the modified segments are spliced in random order and resulting an offspring ACDBE. \subref{fig:Crossover} In crossover,
            two sequences (Parent A and Parent B) are picked from the offspring from
            the mutation process. A sub-sequence of one of the two sequences is randomly
            selected (Sub-sequence A in this case).  Next, random inversion is performed on Sub-sequence A and the
            resulting segment is randomly placed inside an empty sequence of the
            offspring. Lastly, the remaining elements of the offspring are filled by
            the order of the other sequence (Parent B) except the elements that are already in the offspring sequence.}}
    \squeezeup
\end{figure}

\textit{1) Mutation:}
There are three steps for each mutation process, as described in
Fig.~\ref{fig:Mutation}. First, the ordered sequence $\Scal_\text{ECI}$ from the
enhanced cheapest insertion is randomly divided into $k$ segments. Second, random
inversion\footnote{A random inversion of a segment is reversing the order of
destinations in the segment.} is executed for each segment except the first and last
segments, which contain the source and the target. Lastly, the segments in the middle
are randomly reordered and spliced together. Let the resulting offspring sequence be $\psi = \{v_s, p_1,
p_2,\cdots, p_r, v_t\}$, where $v_s$ is the source node, $v_t$ is the target node,
$(r+2)$ is the number of destinations in the sequence (the same cardinality of
$S_\text{ECI}$), and $\{p_i\}_{i=1}^r$ is the re-ordered destinations, which could
possibly contain the start and end. The cost $\Theta$ of the offspring is computed as
\begin{equation}
\label{eq:MulationCost}
    \Theta = \theta(v_s, p_1) + \sum_{i=1}^r \theta(p_i, p_{i+1}) + \theta(p_n, v_t),
\end{equation}
where $\theta(\cdot, \cdot)$ is the path distance between two destinations provided
by the \imomdN.


We perform the mutation process thousands of times, resulting in thousands of offspring. Note that only the offspring with a lower cost than the cost of the parent are kept for the crossover process.

\textit{2) Crossover:}
Let the set of mutated sequences from the mutation process be
$\Psi=\{\psi_i\}_{i=1}^{h}$, where $h$ is the number of offspring kept after the
mutation process. For each generation, the crossover process is performed thousands
of times and only the offspring with a lower cost than the previous generation are
kept. Each crossover process combines sub-sequences of any two sequences ($\psi_i,
\psi_j\in\Psi$) to generate a new offspring, as described in Fig.\ref{fig:Crossover}.
\conference{ The probability of a sequence $\psi_i$ being picked is defined as: 
\begin{equation}
    P_{\psi}(\psi = \psi_i) = \frac{\rho_i}{\sum_{i=1}^{w}\rho_i}, ~~ \rho_i = \frac{1}{\Theta_i},
\end{equation}
where $w$ is the number of the remaining offspring from
each generation after the $(i-1)^{\text{th}}$ generation and $\rho_i$ is the fitness
of the sequence $\psi_i$.
}
Given the two selected sequences and an empty to-be-filled offspring, a segment of
one of the two sequences is randomly selected, and random inversion is performed on
the segment. The resulting segment is randomly placed inside the empty sequence of
the offspring. Lastly, the remaining elements of the offspring are filled by the
order of the other sequence except the elements that are already in the offspring
sequence. After a few generations, the offspring with the lowest cost, $\psi^*$, is the final sequence $\Scal_\text{ECI-Gen}$ of the destinations.

\begin{remark}
    Whenever the IMOMD-RRT$^\ast$ provides a better path (due to its asymptotic optimality), the \tspsolver will be executed to solve for a better visiting order of the destinations. Therefore, the full system provides paths with monotonically improving path cost in an anytime fashion.
\end{remark}

\subsection{Discussion of time complexity}
As mentioned in Sec.~\ref{sec:WIBFS}, the first sequence $\Scal_{\text{current}}$ is constructed by Dijkstra's algorithm, which is an $O(N^2)$ process, where $N$ is the cardinality of the destination set $\Dcal$. We then pass the sequence to the enhanced cheapest insertion algorithm, whose time complexity is $O(N^3)$, to generate a set of parents for the genetic algorithm, which is also an $O(N^3)$ process. Therefore, the overall time complexity of the proposed \tspsolver is $O(N^3)$, and indeed a polynomial solver.

\comment{
Let the fitness $\rho_i$ of the path $\psi_i$ be the inverse of the cost $\Theta_i$ defined in \eqref{eq:MulationCost}: 
\begin{equation}
    \rho_i = \frac{1}{\Theta_i}.
\end{equation}
{The possibility of a path $\psi_i$ being picked:
\begin{equation}
    P_{\psi}(\psi = \psi_i) = \frac{\rho_i}{\sum_{i=1}^{w}\rho_i},
\end{equation}
where $w$ is the number of the entire remaining offspring from each generation.
}
}





\comment{
\begin{figure}[t]%
    \centering
    \includegraphics[trim=0 0 0 0,clip,width=0.6\columnwidth]{graphics/crossover.png}%
    \caption[]{{Crossover in Genetic Algorithm. Two sequences (Parent A and Parent B) are picked from the offspring from the mutation process. A segment of
one of the two sequences is randomly selected. Next, random inversion is performed on the segment and the resulting segment is randomly placed inside an empty sequence of the offspring. Lastly, the remaining elements of the offspring are filled by the
order of the other sequence except the elements that are already in the offspring sequence.}}
\label{fig:Crossover}%
\end{figure}
}

\comment{
\section{ACI-Gen Solver Benchmarking}
\label{sec:ACIGenBenchmarking}
We summarize some terminologies from the graph theory~\cite{west2001introduction, bondy1976graph}. If each pair of vertices in the graph is connected by an edge, the graph is called a complete graph; otherwise an incomplete graph. The number of vertices and the number of edges in the graph is called the graph order and the graph size, respectively. 

To benchmark the proposed ACI-Gen solver, we randomly generated $3000$ complete graphs and incomplete graph with order of 5 to 13. For each graph, the edge connecting two vertices is a straight line and consequently, the resulting graph obeys triangle inequality. A brute force algorithm \cite{cormen2022introduction} that permutes all the possible solutions is implemented as a baseline. The brute force algorithm guarantees the optimal path for each graph. For each order of graph, we compute the average length ratio $\rho_\text{mean}$, the standard deviation ratio $\rho_\text{std}$, the ratio of optimality $\rho_\text{optimality}$, and the worst case $\rho_\text{worst}$. In particular, they are computed as 
\begin{align}
    \rho_\text{mean}^o &= \frac{\sum_{i=1}^{3000}{}^i\Theta_\text{Brute-Force}^*}{\sum_{i=1}^{3000}{}^i\Theta_\text{ACI-Gen}^*}\\
    \rho_\text{std}^o &= \mathtt{std}(\frac{{}^i\Theta_\text{Brute-Force}^*}{{}^i\Theta_\text{ACI-Gen}^*})\\
    \rho_\text{optimality}^o &= \frac{\sum_{i=1}^{3000}{{I({}^i\Theta_\text{Brute-Force}^* == {}^i\Theta_\text{ACI-Gen}^*)}}}{3000}\\
    \rho_\text{worst}^o &= \argmax\frac{{}^i\Theta_\text{Brute-Force}^*}{{}^i\Theta_\text{ACI-Gen}^*},
\end{align}
where $o$ is the order of the graph and $I$ is the indicator function returning one if the solution is optimal; otherwise zero.

Table~\ref{tab:aci_gen_complete} and Table~\ref{tab:aci_gen_incomplete} show the results of complete graphs and incomplete graphs, respectively. In fig.~\ref{fig:CompleteTSP} and fig.~\ref{fig:IncompleteTSP}, we  show the results in charts for better visualization.



\begin{table}[t]
\caption{ACI-Gen Performance on Complete Graphs}
\label{tab:aci_gen_complete}
\scalebox{0.9}{
\begin{tabular}{|c|c|c|c|c|}
\hline
\begin{tabular}[c]{@{}c@{}}Graph\\ Order\end{tabular} & \begin{tabular}[c]{@{}c@{}}Average \\ Cost Ratio\end{tabular} & \begin{tabular}[c]{@{}c@{}}Optimal \\ Result Ratio\end{tabular} & \begin{tabular}[c]{@{}c@{}}Worst \\ Cost ratio\end{tabular} & \begin{tabular}[c]{@{}c@{}}STD of \\ Cost Ratio\end{tabular} \\ \hline
5                                                     & 1.000                                                         & 100.000                                                         & 1.000                                                       & 0.000                                                        \\ \hline
6                                                     & 1.000                                                         & 100.000                                                         & 1.000                                                       & 0.000                                                        \\ \hline
7                                                     & 1.000                                                         & 100.000                                                         & 1.000                                                       & 0.000                                                        \\ \hline
8                                                     & 1.000                                                         & 100.000                                                         & 1.000                                                       & 0.000                                                        \\ \hline
9                                                     & 1.000                                                         & 100.000                                                         & 1.000                                                       & 0.000                                                        \\ \hline
10                                                    & 1.000                                                         & 100.000                                                         & 1.000                                                       & 0.000                                                        \\ \hline
11                                                    & 1.000                                                         & 99.967                                                          & 1.001                                                       & 0.000                                                        \\ \hline
12                                                    & 1.000                                                         & 99.867                                                          & 1.035                                                       & 0.001                                                        \\ \hline
13                                                    & 1.000                                                         & 99.267                                                          & 1.034                                                       & 0.001                                                        \\ \hline
\end{tabular}
}
\end{table}

\begin{table}[t]
\caption{ACI-Gen Performance on Incomplete Graphs}
\label{tab:aci_gen_incomplete}
\scalebox{0.9}
{
\begin{tabular}{|c|c|c|c|c|}
\hline
\begin{tabular}[c]{@{}c@{}}Graph\\ Order\end{tabular} & \begin{tabular}[c]{@{}c@{}}Average \\ Cost Ratio\end{tabular} & \begin{tabular}[c]{@{}c@{}}Optimal \\ Result Ratio\end{tabular} & \begin{tabular}[c]{@{}c@{}}Worst \\ Cost ratio\end{tabular} & \begin{tabular}[c]{@{}c@{}}STD of \\ Cost Ratio\end{tabular} \\ \hline
5                                                     & 1.004                                                         & 95.967                                                          & 1.422                                                       & 0.028                                                        \\ \hline
6                                                     & 1.005                                                         & 93.833                                                          & 1.378                                                       & 0.026                                                        \\ \hline
7                                                     & 1.005                                                         & 91.767                                                          & 1.489                                                       & 0.028                                                        \\ \hline
8                                                     & 1.006                                                         & 89.967                                                          & 1.251                                                       & 0.024                                                        \\ \hline
9                                                     & 1.007                                                         & 88.400                                                          & 1.288                                                       & 0.024                                                        \\ \hline
10                                                    & 1.006                                                         & 87.067                                                          & 1.378                                                       & 0.025                                                        \\ \hline
11                                                    & 1.008                                                         & 83.533                                                          & 1.380                                                       & 0.024                                                        \\ \hline
12                                                    & 1.008                                                         & 81.667                                                          & 1.332                                                       & 0.023                                                        \\ \hline
13                                                    & 1.015                                                         & 74.906                                                          & 1.375                                                       & 0.026                                                        \\ \hline
\end{tabular}
}
\end{table}

\begin{figure}[t]%
    \centering
    \subfloat{%
    \includegraphics[trim=0 0 0 0,clip,width=0.48\columnwidth]{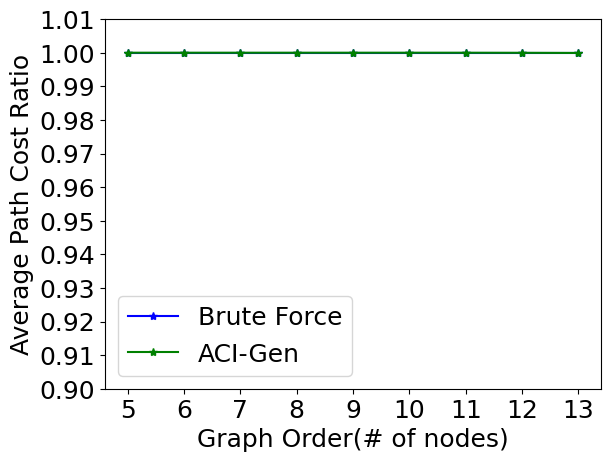}}~
    \subfloat{%
    \includegraphics[trim=0 0 0 0,clip,width=0.48\columnwidth]{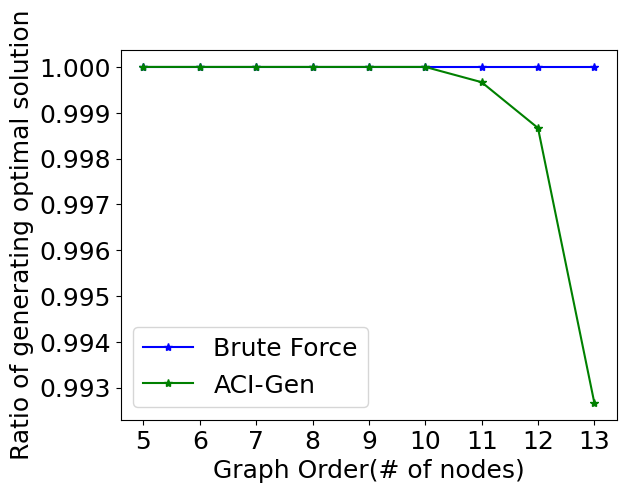}}\\
    \subfloat{%
    \includegraphics[trim=0 0 0 0,clip,width=0.48\columnwidth]{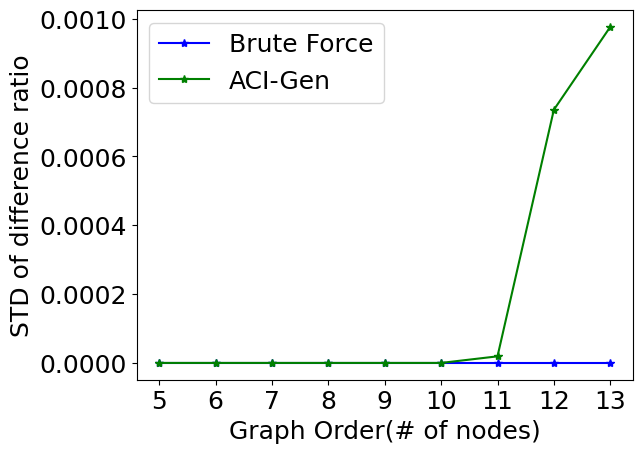}}~
    \subfloat{%
    \includegraphics[trim=0 0 0 0,clip,width=0.48\columnwidth]{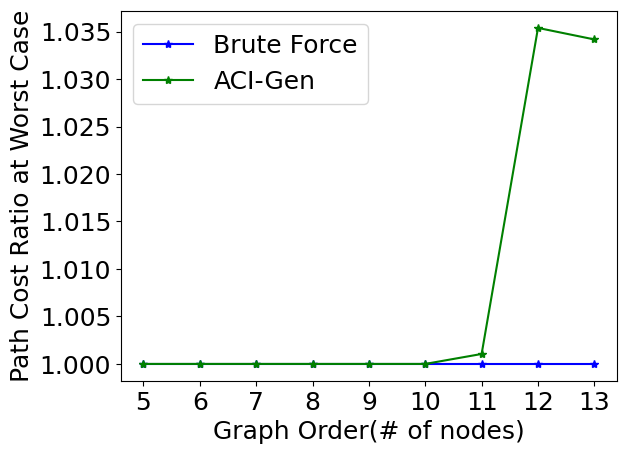}}%
    \caption[]{\bdd{The validation of the \tspsolver on complete graphs.}
    }%
    \label{fig:CompleteTSP}%
\end{figure}

\begin{figure}[t]%
    \centering
    \subfloat{%
    \includegraphics[trim=0 0 0 0,clip,width=0.48\columnwidth]{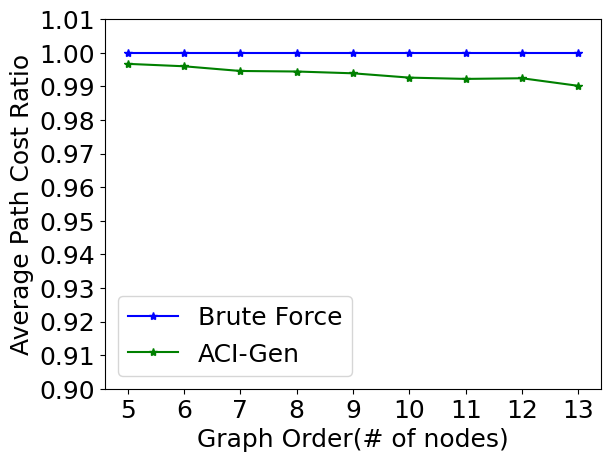}}~
    \subfloat{%
    \includegraphics[trim=0 0 0 0,clip,width=0.48\columnwidth]{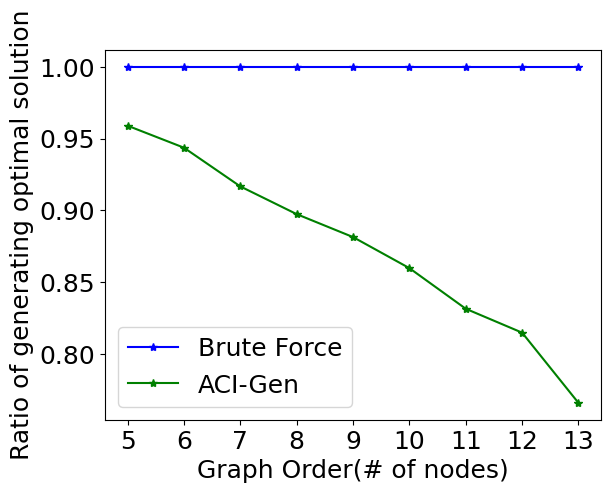}}\\
    \subfloat{%
    \includegraphics[trim=0 0 0 0,clip,width=0.48\columnwidth]{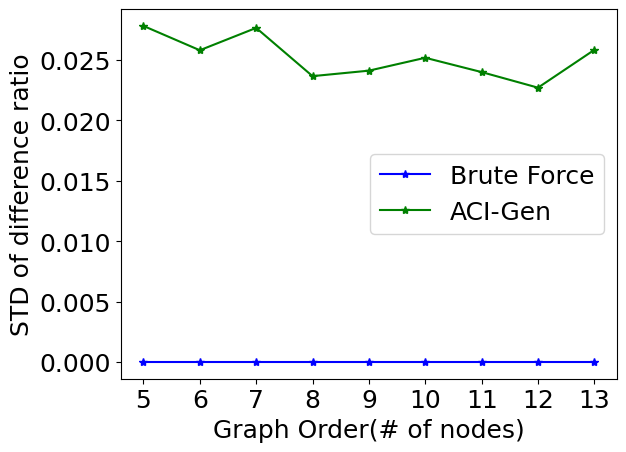}}~
    \subfloat{%
    \includegraphics[trim=0 0 0 0,clip,width=0.48\columnwidth]{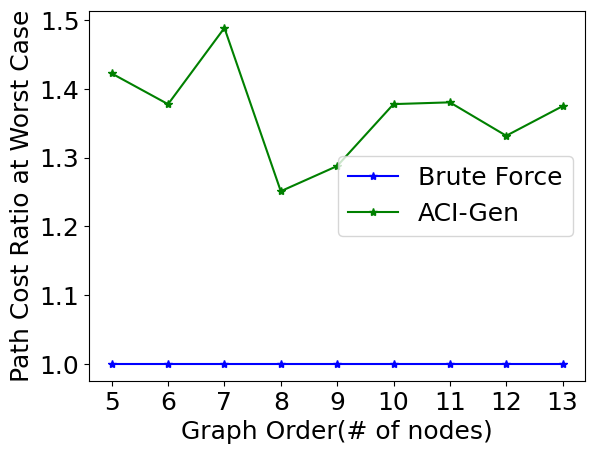}}%
    \caption[]{\bdd{The validation of the \tspsolver on incomplete graphs.}
    }%
    \label{fig:IncompleteTSP}%
\end{figure}
}
\section{Experimental Results}
\label{sec:Experiments}
\conference{
This section presents extensive evaluations of the \imomt system applied to two complex vehicle routing scenarios.
The robot state $\zeta$ is defined as latitude and longitude. The distance between robot states $\mathtt{Dist}(\cdot, \cdot)$ in \eqref{eq:NearestNode} is defined as the haversine distance\cite{van2012heavenly}. We implemented the
bi-directional A$^*$ \cite{BiAstar} and ANA$^*$ \cite{ANAstar} as our baselines to
compare the speed and memory usage (the number of explored nodes). We evaluate the IMOMD-RRT$^\ast$
system (IMOMD-RRT$^\ast$ and the \tspsolverN) on a large and complex map of Seattle, USA. The map contains $1,054,372$ nodes
and $1,173,514$ edges, and is downloaded from OpenStreetMap (OSM), which is a public
map service built for real applications\footnote{Apple Map$^\text{\textregistered}$
actually uses OpenStreetMap as their foundation.}. We then place $25$ destinations in the map.
We demonstrate that the \imomt system is able to concurrently find paths connecting destinations and determine the order of
destinations. We also show that the system escapes from a bug trap by inherently receiving prior knowledge. The algorithm runs on a laptop
equipped with Intel$^\text{\textregistered}$ Core$^{\text{TM}}$ i7-1185G7 CPU @ 3.00 GHz.}

\comment{
This section presents extensive evaluations of the \imomt system. We also implemented
bi-directional \Astar and ANA$^*$ as two baseline to compare the speed and the number
of explored nodes. We chose to run the \imomt system on complex graph-based maps built from
real data. The algorithm runs on a laptop equipped with
\bdd{Intel$^\text{\textregistered}$ Core$^{\text{TM}}$ i7-9750H CPU @ 2.60 GHz.} We
demonstrate the \imomt system is able to escape from bug traps, and to receive prior
knowledge, as well as to concurrently find paths connecting destinations and
determine the order of destinations.
}

\begin{figure}[t]
    \centering
    \subfloat{%
    \includegraphics[trim=0 0 0 0,clip,height=0.22\columnwidth]{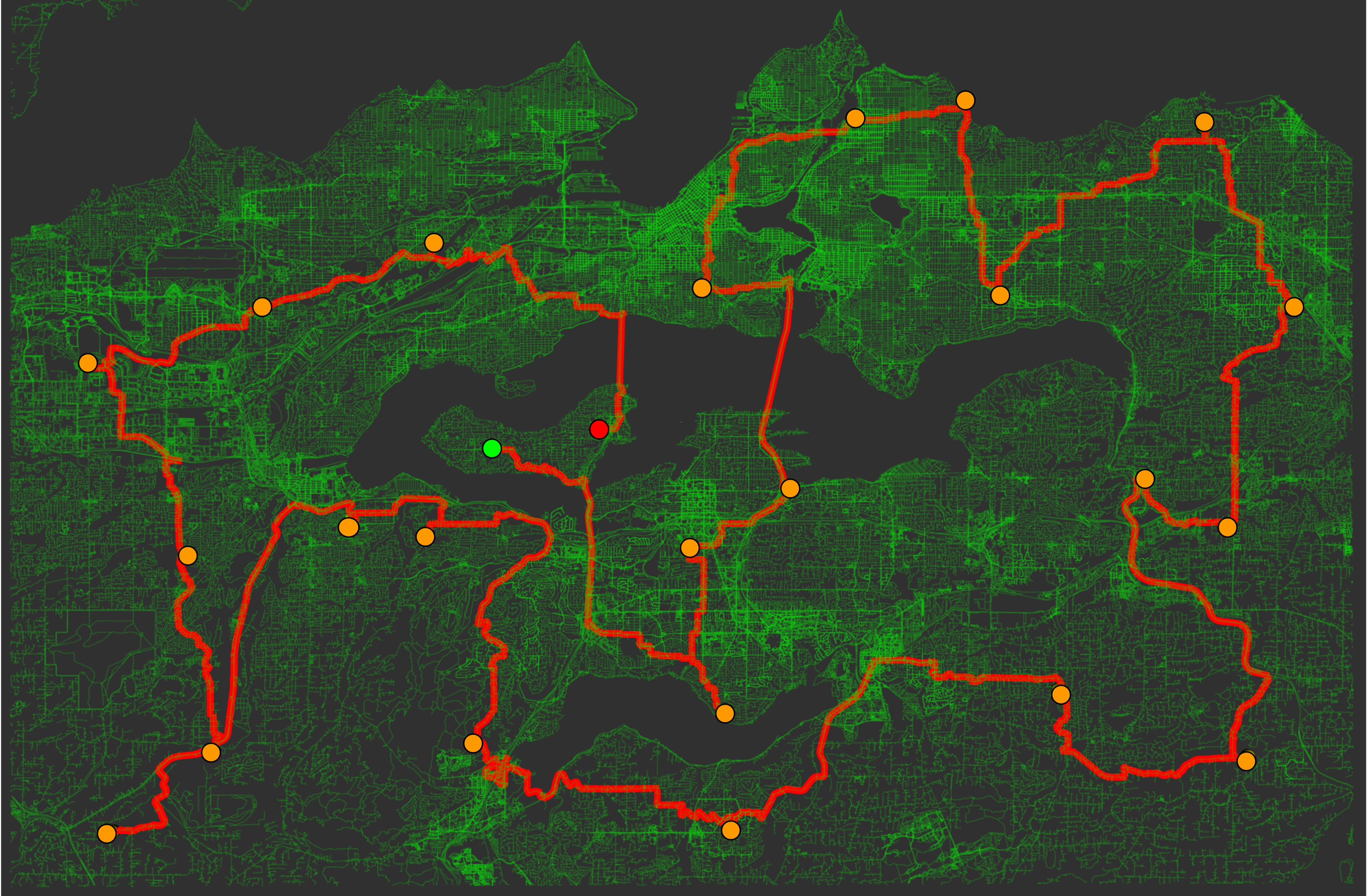}}~~
    \subfloat{%
    \includegraphics[trim=0 0 0 130,clip,height=0.22\columnwidth]{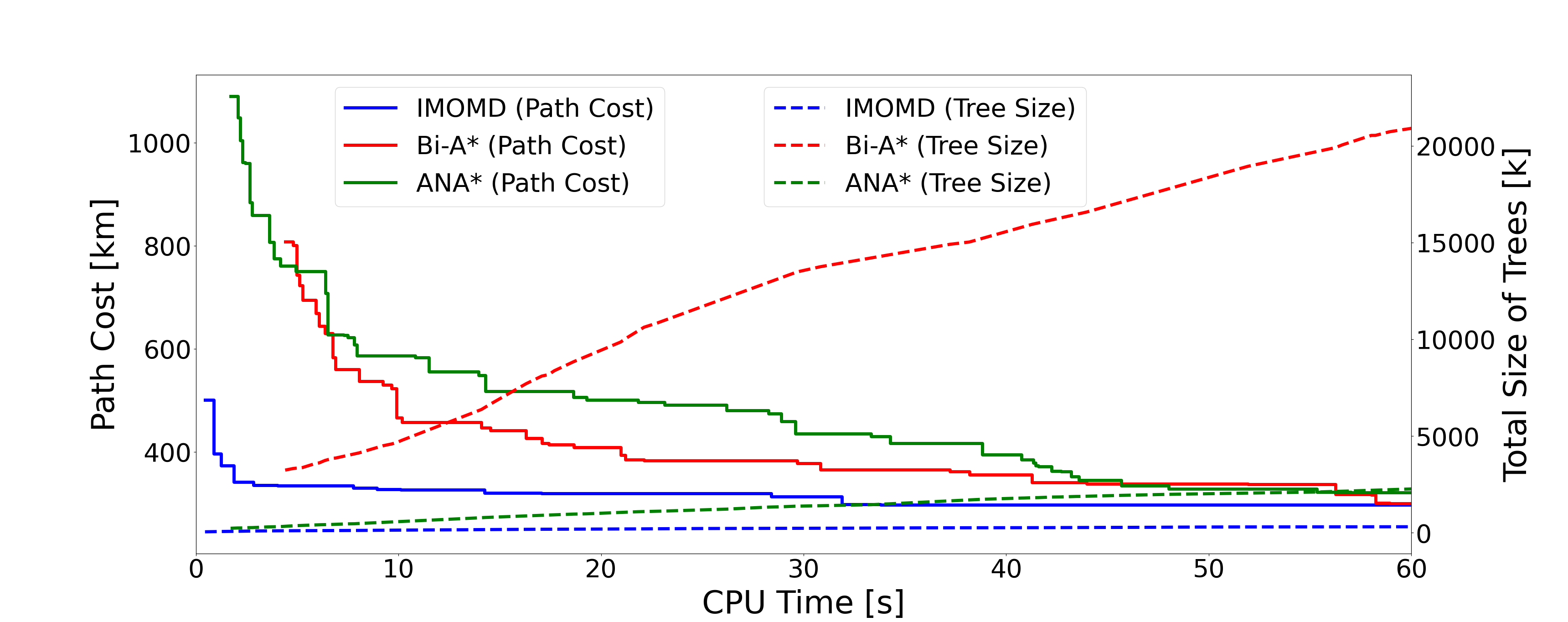}}
    \caption[]{{Quantitative and qualitative results for an OSM of Seattle, where we have 25 destinations to be visited. The proposed \imomd outperforms Bi-A$^\ast$ and ANA$^\ast$ in term of speed and memory usage (the number of explored nodes).}
    }%
    \label{fig:OSMSeattle}%
    \squeezeup
\end{figure}

\comment{
\subsection{Open Street Map: Bug Traps}
\label{sec:BugTraps}
}
\conference{
To show the performance and ability of multi-objective and
determining the visiting order, we randomly set $25$ destinations in the Seattle map. There are 25! possible combinations of visiting orders and therefore it is intractable to solve the visiting order by brute force.
The results are shown in Fig.~\ref{fig:OSMSeattle}, where \imomd finds the first path
faster than both Bi-A$^\ast$ and ANA$^\ast$ with a lower cost and then also spends less time
between solution improvements. Additionally, the memory usage of \imomd is less than
ANA* and much less than bi-A$^\ast$. As shown in Table~\ref{tab:ExpResults}, the proposed system provides the first solution 10 times faster than bi-A$^\ast$ and four time faster than ANA$^\ast$. In addition, the proposed system also consumes 65 times less memory than bi-A$^\ast$ and 4.7 times less memory usage than ANA$^\ast$.



\begin{table}[t]
\caption{Quantitative results of the proposed \imomd system on two large maps (both graphs contain more than one million nodes and edges) built for real robotics and vehicle applications. The proposed system outperforms bi-A$^\ast$ and ANA$^\ast$.}
\label{tab:ExpResults}
\centering
\begin{tabular}{|c|c|r|r|r|}
\hline
 &
   &
  \begin{tabular}[c]{@{}c@{}}Initial Solution Time \\ {[}seconds{]}\end{tabular} &
  \begin{tabular}[c]{@{}c@{}}Initial Path Cost \\ {[}kilometers{]}\end{tabular} &
  \begin{tabular}[c]{@{}c@{}}Final Memory Usage \\ {[}\# explored nodes{]}\end{tabular} \\ \hline
\multirow{3}{*}{Seattle}       & IMOMD-RRT$^\ast$ & \textbf{0.44} & \textbf{501,342} & \textbf{49,768} \\ \cline{2-5} 
                               & Bi-A$^\ast$ & 4.40          & 808,416          & 3,240,515       \\ \cline{2-5} 
                               & ANA$^\ast$  & 1.70          & 1,089,873        & 234,457         \\ \hline
\multirow{3}{*}{San Francisco} & IMOMD-RRT$^\ast$ & \textbf{1.10} & \textbf{156,807} & \textbf{61,785} \\ \cline{2-5} 
                               & Bi-A$^\ast$ & 9.93          & 315,061        & 3,640,863       \\ \cline{2-5} 
                               & ANA$^\ast$  & Failed            & Failed               & Failed              \\ \hline
\end{tabular}
\end{table}

Prior knowledge through pseudo
destinations can also be leveraged to traverse challenging topology, such as bug-traps\cite{Lav06}.
This problem
is commonly seen in man-made environments such as a neighborhood with a single entry
or cities separated by a body of water, as in Fig.~\ref{fig:InformBugTrapOSM}. As mentioned in Remark~\ref{sec:Informable}, prior knowledge is provided as
a number of pseudo
destinations in the \imomt as a prior collision-free path in the graph
for robotics inspection or vehicle routing. Next, the prior path is then being
rewired by the \imomt to improve the path. We demonstrate this feature by providing
the prior knowledge to escape the bug trap in San Francisco, as shown in Fig.~\ref{fig:InformBugTrapOSM}. The map contains $1,277,702$ nodes and $1,437,713$ edges. As shown in Table~\ref{tab:ExpResults}, the proposed system escapes from the trap nine times faster than bi-A$^\ast$, whereas ANA$^\ast$ failed to provide a path within the given time frame. The proposed system also consumes 58.9 times less memory than bi-A$^\ast$.
}

In summary, we developed an anytime iterative system to provide paths between multiple objectives by the \imomd and to determine the visiting order of the objectives by the \tspsolver solver in polynomial time. We also demonstrate that the proposed system is able to inherently accommodate prior information to escape from challenging topology.

\conference{
\begin{figure}[t]%
    \centering
    \subfloat{%
    \includegraphics[trim=0 0 0 0,clip,height=0.22\columnwidth]{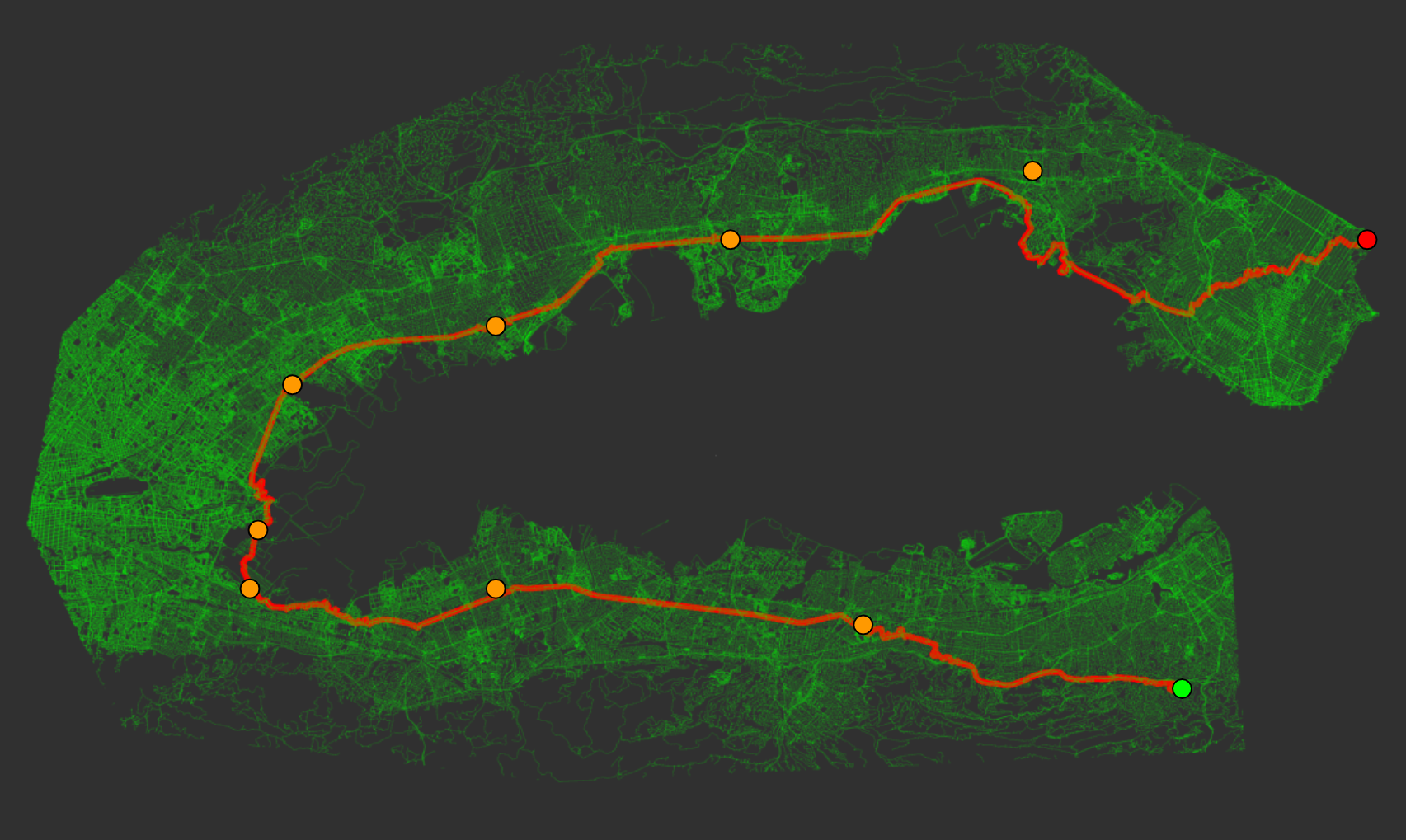}}
    \subfloat{%
    \includegraphics[trim=0 0 0 100,clip,height=0.22\columnwidth]{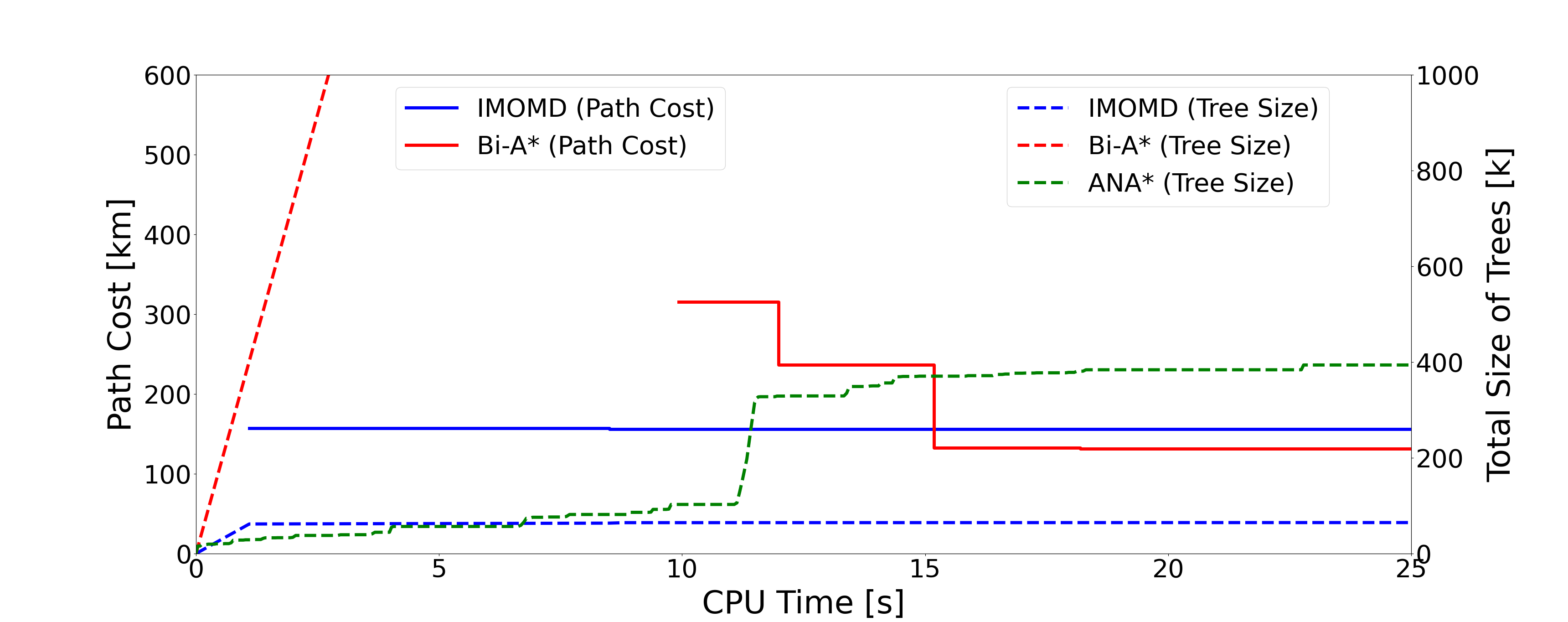}}~
    \caption[]{{Providing prior knowledge to the proposed \imomd system to avoid bug traps. The left and the right are the qualitative and quantitative results for a bug trap in San Francisco, respectively. We have eight pseudo destinations to help escape the challenging topology, where the source and target are separated by a body of water. Note that ANA$^\ast$ failed to provide a solution in the given time.}
    }%
    \label{fig:InformBugTrapOSM}%
\end{figure}
}

\section{Conclusion and Future Work}
\label{sec:Conclusion}
\conference{
We presented an anytime iterative system on large-complex graphs to solve the
multi-objective path planning problem, to decide the visiting order of the
objectives, and to incorporate prior knowledge of the potential trajectory. The
system is comprised of an anytime informable multi-objective and multi-directional
RRT$^*$ to connect the destinations to form a connected graph and the ECI-Gen solver
to determine the visiting order (via a relaxed Traveling Salesman Problem) in polynomial
time. 

The system was extensively evaluated on OpenStreetMap (OSM), built for autonomous
vehicles and robots in practice. In particular, the system solved a path planning
problem and the visiting order with $25$ destinations ($25!$ possible combinations of visiting orders) on an OSM of Seattle, containing more than a million nodes and edges, in
0.44 seconds. In addition, we demonstrated the system is able to leverage a reference path (prior knowledge) to navigate challenging topology for robotics
inspection or vehicle routing applications. All the evaluations show that our proposed method outperforms
the Bi-A$^\ast$ and ANA$^\ast$ algorithm in terms of speed and memory usage. 
}

\comment{
We presented an anytime iterative system on large-complex graphs to solve the
multi-objective path planning problem, to decide the visiting order of the
objectives, and to incorporate prior knowledge of the potential path. The
system comprised of an anytime informable multi-objective and multi-directional
RRT$^*$ to connect the destinations to form a union graph and a weighted-insertion
breadth-first search algorithm solver to determine the visiting order in polynomial
time. We implemented informable bi-directional A$^*$, bi-directional A$^*$, and
ANA$^*$ as our baselines. The system was extensively evaluated on large-complex
graphs built for autonomous vehicles and robots.
In particular, we demonstrated the system is able to escape from bug
traps in the graphs, and showed that the prior knowledge of robotics inspection via
providing prior knowledge to the bug-trap experiments and the system converged much
faster. We also ran the proposed system on maps of San Francisco, New York, and
Chicago from OpenStreetMap. 
\comment{
To further evaluate the system, we examine the system
on maps built for autonomous vehicles from Toyota Research Institute. 
}
}

In the future, we shall use the developed system within autonomy
systems\cite{rehder2016extending, furgale2013unified, oth2013rolling,
huang2020improvements, huang2021lidartag, huang2021optimal, huang2020intinsic,
Hartley-RSS-18, hartley2019contact, huang2021efficient,
gong2021zero, gong2020angular, gong2019feedback} on a robot to perform point-to-point
tomometric navigation in graph-based maps while locally avoiding obstacles and uneven
terrain. It would also be interesting to deploy the system with
multi-layered graphs and maps \cite{Fankhauser2018ProbabilisticTerrainMapping, Fankhauser2014RobotCentricElevationMapping, Lu2020BKI} to incorporate different types of information.



\section*{Acknowledgment}
\small{
The first author conceptualized and initiated the research problem, designed the
    components of the system, determined the evaluation metrics, interpreted the
    results, and led the project. The first, third, fourth, and last author wrote this
    manuscript. The second author consolidated the initial version of the enhanced
    cheapest insertion algorithm, and implemented the initial version of the system.
    The third author helped conceptualize the entire current \imomd system including
    the \imomt and the ECI-Gen solver, provided perceptive literature review,
    implemented the components of the system and all the baselines, and ran the
    system on various of maps. \comment{The third author benchmarked the ECI-Gen
    solver by evaluating it on a number of orders of graphs.} The fourth and the last
    author provided insightful knowledge to the full system, suggested practical
improvement to the system, and guided the direction of the project as well as
supported the work. The first author would like to thank all the authors for
assisting the research and for all of the conversations. Toyota Research Institute
provided funds to support this work. Funding for J. Grizzle was in part provided by
NSF Award No.~2118818. This article solely reflects the opinions and conclusions of
its authors and not the funding entities.The first author
thanks Wonhui Kim for useful conversations.
} 


\bibliographystyle{DefinesBib/bib_all/IEEEtran}
{ \small
\bibliography{DefinesBib/bib_all/strings-abrv,DefinesBib/bib_all/ieee-abrv,DefinesBib/bib_all/BipedLab.bib,DefinesBib/bib_all/Books.bib,DefinesBib/bib_all/Bruce.bib,DefinesBib/bib_all/ComputerVision.bib,DefinesBib/bib_all/ComputerVisionNN.bib,DefinesBib/bib_all/IntrinsicCal.bib,DefinesBib/bib_all/L2C.bib,DefinesBib/bib_all/LibsNSoftwares.bib,DefinesBib/bib_all/ML.bib,DefinesBib/bib_all/OptimizationNMath.bib,DefinesBib/bib_all/Other.bib,DefinesBib/bib_all/StateEstimationSLAM.bib,DefinesBib/bib_all/MotionPlanning.bib,DefinesBib/bib_all/Mapping.bib,DefinesBib/bib_all/TrajectoriesOptimization.bib,DefinesBib/bib_all/GraphTheory.bib}
}

\end{document}